\def\year{2021}\relax
\documentclass[letterpaper]{article} 
\usepackage{aaai21}  
\usepackage{times}  
\usepackage{helvet} 
\usepackage{courier}  
\usepackage[hyphens]{url}  
\usepackage{graphicx} 
\usepackage{natbib}  
\usepackage{caption} 
\usepackage{amsmath}
\usepackage{amssymb}
\usepackage{subfigure}
\usepackage{color}
\usepackage{epsfig}
\usepackage{epstopdf}
\usepackage{algorithm}
\usepackage{algorithmic}
\usepackage{multirow}
\usepackage{comment}
\usepackage{diagbox}

\usepackage[switch]{lineno}

\title{Investigate Indistinguishable Points in Semantic Segmentation of 3D Point Cloud}

\author{
    Mingye Xu\textsuperscript{\rm 1, \rm 2}\thanks{M.Xu and Z.Zhou contributed equally.},
    Zhipeng Zhou\textsuperscript{\rm 1, \rm 4}\footnotemark[1],
    Junhao Zhang\textsuperscript{\rm 1},
    Yu Qiao\textsuperscript{\rm 1 \rm 3}\thanks{Corresponding author.} \\ 
}

\affiliations{
	\textsuperscript{\rm 1}ShenZhen Key Lab of Computer Vision and Pattern Recognition,\\ SIAT-SenseTime Joint Lab,Shenzhen Institutes of Advanced Technology, Chinese Academy of Sciences\\ 
	\textsuperscript{\rm 2}University of Chinese Academy of Sciences, China\\ 
	\textsuperscript{\rm 3}Shanghai AI Lab, Shanghai, China\\
	\textsuperscript{\rm 4}SIAT Branch, Shenzhen Institute of Artificial Intelligence and Robotics for Society\\
	\{my.xu, zp.zhou, zhangjh, yu.qiao\}@siat.ac.cn 
}
\begin{document}
\maketitle

\begin{abstract}

This paper investigates the indistinguishable points (difficult to predict label) in semantic segmentation for large-scale 3D point clouds. The indistinguishable points consist of those located in complex boundary, points with similar local textures but different categories, and points in isolate small hard areas, which largely harm the performance of 3D semantic segmentation. To address this challenge, we propose a novel Indistinguishable Area Focalization Network (IAF-Net), which selects indistinguishable points adaptively by utilizing the hierarchical semantic features and enhance fine-grained features for points especially those indistinguishable points. We also introduce multi-stage loss to improve the feature representation in a progressive way. Moreover, in order to analyze the segmentation performances of indistinguishable areas, we propose a new evaluation metric called Indistinguishable Points Based Metric (IPBM). Our IAF-Net achieves the comparable results with state-of-the-art performance on several popular 3D point cloud datasets e.g. S3DIS and ScanNet, and clearly outperforms other methods on IPBM. Our code will be available at \url{https://github.com/MingyeXu/IAF-Net}
\end{abstract}
\section{Introduction}

Deep learning on point cloud analysis has been attracting more and more attention recently. Among the tasks of point cloud analysis, efficient semantic segmentation of large-scale 3D point cloud is a challenging task with huge applications \cite{rusu2008towards, chen2017multi, 8805456}.
A key challenge is that 3D point cloud semantic segmentation relies on unstructured data which is typically irregularly sampled and unordered. 
Due to the complexity of large-scale 3D point cloud, this task also requires the understanding of the fine-grained details for each point.

\begin{figure}[t]
\centering
\includegraphics[width=1\columnwidth]{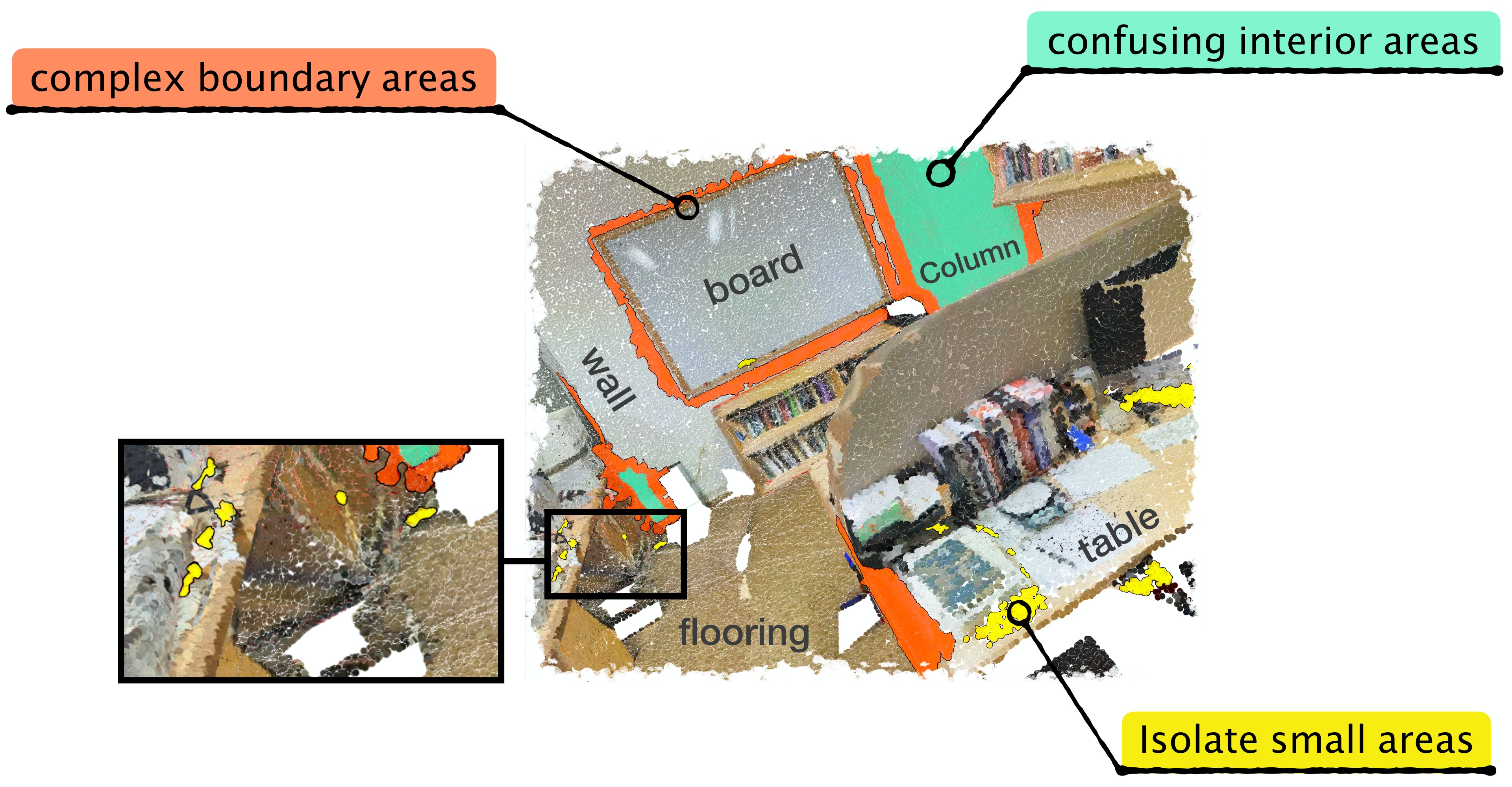} 
\caption{Three types of the indistinguishable areas.}
\label{fig_introduction_2}
\end{figure}

For point cloud semantic segmentation, there exist some areas which are hard to be segmented, and we name these areas as “indistinguishable” areas. In order to analyze the image semantic segmentation results in detail, \cite{li2017not} divide pixels into different difficulty levels. 
Inspired by \cite{li2017not}, we can also categorize these “indistinguishable” areas into three types (Figure \ref{fig_introduction_2}): 
The first type is called \textbf{complex boundary areas} (orange areas in Figure \ref{fig_introduction_2}) which belong to the boundary points (object boundaries and prediction boundaries).
In most cases, it is difficult to identify the boundaries between different objects accurately. Because the features of each point are characterized by the information of local regions, the predictions of the boundary points will be over smooth between objects of different categories which are close in Euclidean space. 
The second type is named \textbf{confusing interior areas} (cyan areas in Figure \ref{fig_introduction_2}) which contain interior points from objects of different categories with similar textures and geometric structures. For example, door and wall have similar appearance which are almost flat and share similar colors. Even for human being, it is hard to identify part of door and wall accurately in these cases. 
The last type is called \textbf{isolate small areas} (yellow areas in Figure \ref{fig_introduction_2}), which are scattered and hard to be predicted.
In addition, objects in the scenes would not be fully captured by the devices because of the occlusion. All of the challenges mentioned above hinder the accuracy of semantic segmentation of 3D point cloud.
As far as we know, these “indistinguishable” points are not deeply explored in most of the previous methods \cite{jiang2018pointsift, yang2019modeling} on point cloud semantic segmentation task.

To improve the segmentation performance on indistinguishable points mentioned above, we design an efficient neural network which is able to enhance the features of points especially indistinguishable points. 
However, this task has two challenges to be addressed: 1) How to discover indistinguishable points adaptively in the training process; 2) How to enhance the features of these points. To this end, we propose a new module called Indistinguishable Areas Focalization (IAF) which can adaptively select indistinguishable points considering hierarchical semantic features.
To enhance the features of indistinguishable points, IAF module firstly acquires the fine-grained features and high-level semantic features of indistinguishable points, then enhances the features through an nonlocal operation between these points and the corresponding whole point set.
Furthermore, we introduce a multi-stage loss function $L_{ms}$ to strengthen the feature descriptions of corresponding points in each layer. In this way, we can capture the features in a more progressive manner, which guarantees the accuracy of features in each layer.

\begin{figure}[t]
\centering
\includegraphics[width=1\columnwidth]{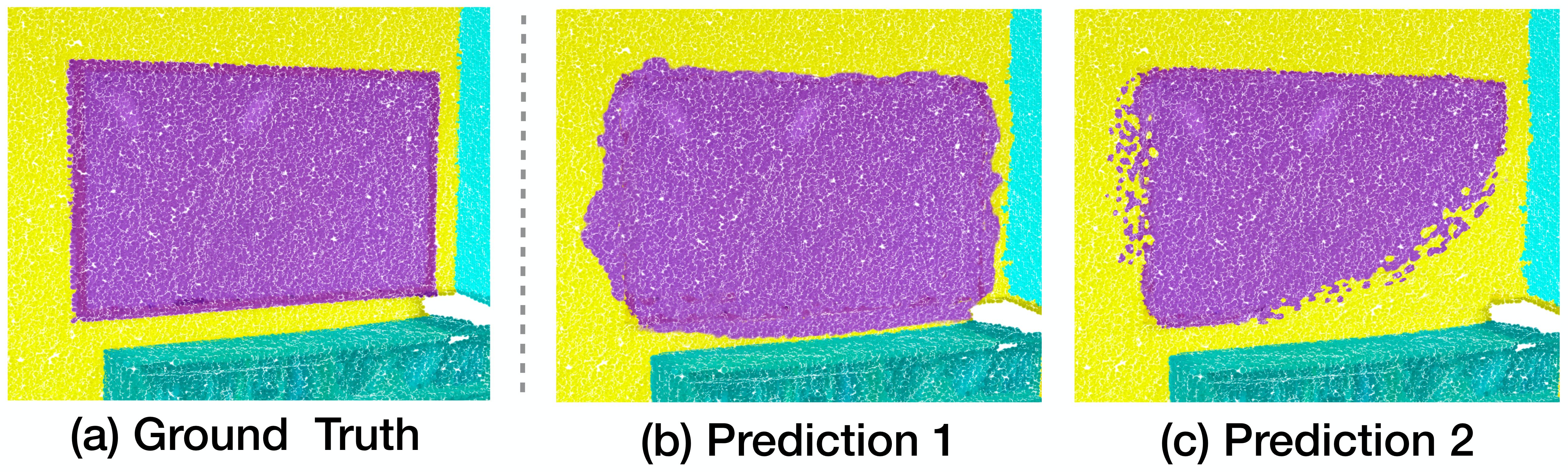} 
\caption{(a) is the ground truth of a room. (b) and (c) are two different predictions which result in similar m-IoU.}
\label{fig_introduction_3}
\end{figure}

Mean IoU (m-IoU) and Overall Accuracy (OA) are two widely-used evaluation metrics of 3D semantic segmentation. 
OA describes the average degree of accuracy which ignores the various distribution of different categories of objects. m-IoU can reflect the accuracy of model on the identification of each category independently. 
Under certain circumstances (Figure \ref{fig_introduction_3} shows), the visualizations of two predictions with similar m-IoU can be totally different in details.
In order to cooperate with the indistinguishable points' partitions and to provide a supplementary metric for OA and m-IoU,
we propose a novel evaluation metric named Indistinguishable Points Based Metric (IPBM). 
This evaluation metric focuses on different types of indistinguishable areas. With this evaluation metric, we can evaluate the effectiveness of segmentation methods more objectively and more granularly. It has a certain contribution to the segmentation task evaluation in the future.

The main contributions are summarized as follows,
\begin{itemize}
\item We propose the Indistinguishable Areas Focalization (IAF) module
which can select indistinguishable points adaptively and enhance the features of each point.
\item  We utilize the multi-stage loss to strengthen feature descriptions in each layer, which guarantees the features can represent points more accurately in a progressive way. 
\item Our method achieves the comparable performances with state-of-the-art methods on several popular datasets for 3D point cloud semantic segmentation.
\item We introduce Indistinguishable Points Based Metric (IPBM) which focuses on the performances of segmentation methods on different types of  indistinguishable areas. 
\end{itemize}

\section{Related Work}
\textbf{Point-Based Networks.}
Point-based networks work on irregular point clouds directly. 
Inspired by PointNet \cite{Charles_2017} and PointNet++ \cite{qi2017pointnet++}, many recent works \cite{Hu_2020, Han_2020, zhang-shellnet-iccv19, Xu_2020,wu2019pointconv} propose different kinds of modules based on pointwise MLP.
ShellNet \cite{zhang-shellnet-iccv19} introduce ShellConv which can allow efficient neighbor point query simultaneously and resolve point order ambiguity by defining a convolution order from inner to the outer shells on a concentric spherical domain.
RandLANet \cite{Hu_2020} utilize a local feature aggregation module to automatically preserve complex local structures by progressively increasing the receptive field for each point. Some works construct novel and efficient point convolutions. A-CNN \cite{Komarichev_2019} propose a multi-level hierarchical annular convolution which can set arbitrary kernel sizes on each local ring-shaped domain to better capture shape details. KPConv \cite{thomas2019kpconv} apply a new convolution based operator which uses a set of kernel points to define the area where each kernel weight is applied. 
However, most methods do not consider the indistinguishable points on point cloud semantic segmentation task specially. 
By contrast, we propose a novel IAF module which enhances the features of points, especially points in the indistinguishable areas. IAF module in each layer uses a specially designed points selection operation to mine the indistinguishable points adaptively and applies the nonlocal operation to fuse the features between indistinguishable points and corresponding whole point set in each layer. Multi-stage loss is conducive to abstracting representative features in a progressive way.

\begin{figure*}[t]
\centering
\includegraphics[width=1.8\columnwidth]{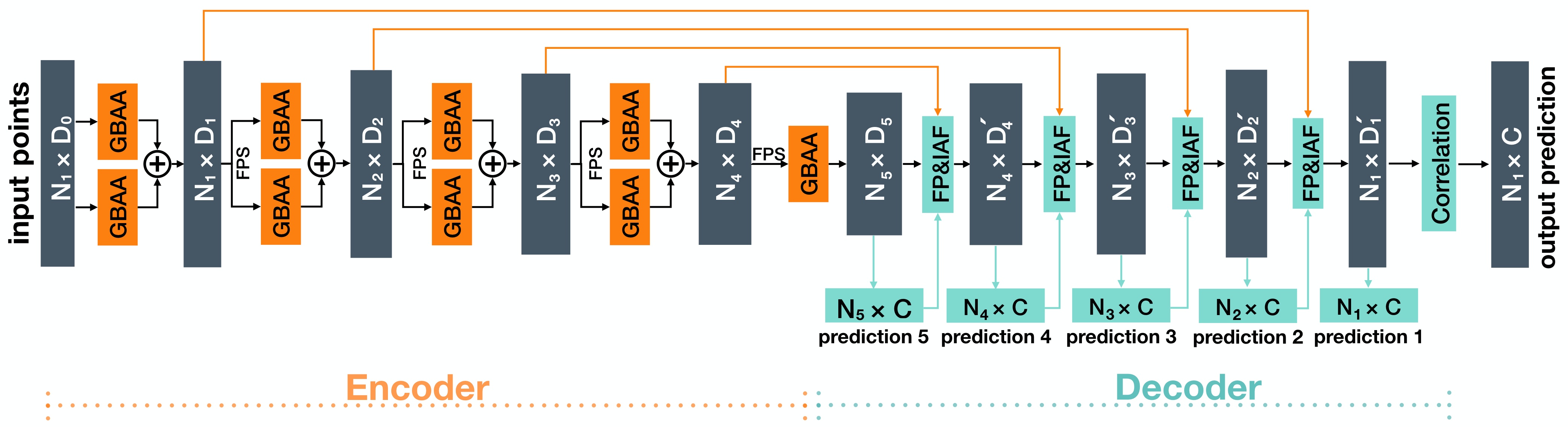} 
\caption{The detailed architecture of our IAF-Net. $(N,D)$ represents the number of points and feature dimension respectively. FPS: Farthest Point Sampling. }
\label{fig_method_network}
\end{figure*}

\noindent\textbf{Local-Nonlocal Mechanism.}
Nonlocal mechanism has been applied to various tasks of computer vision \cite{yan2020pointasnl, cao2019gcnet}. The pioneer work Nonlocal \cite{Wang_2018} in video classification present non-local operations as an efficient, simple and generic component for capturing long range dependencies with deep neural networks. It computes the response at a position as a weighted sum of the features at all positions in the input feature maps. Point2Node \cite{Han_2020} utilize both local and non-local operations to dynamically explore the correlation among all graph nodes from different levels and adaptively aggregate the learned features. Local correlation and non-local correlation are used in a serial way which largely enhances nodes characteristic from different scale correlation learning. To apply the local and non-local mechanism in a targeted way, we use the non-local operation to fuse the features of different layers, which help to enhance the features of indistinguishable points. Moreover, the local features in our network are enhanced by using the multi-stage loss progressively.

\section{Method}\label{Sec_methods}
We denote the point cloud as $\mathrm{P}=\{p_i \in R^{3+d}, i=1,2,...,N\}$ , where $N$ is the number of points and $3+d$ denotes the xyz-dimension and additional properties, such as colors, normal vectors. Like previous works, our IAF-Net mainly consists of encoder and decoder modules which are shown in Figure \ref{fig_method_network}. We will show the details of each part in the following subsections.

\subsection{Encoder Module: Geometry Based Attentive Aggregation}
\begin{figure}[t]
\centering
\includegraphics[width=1\columnwidth]{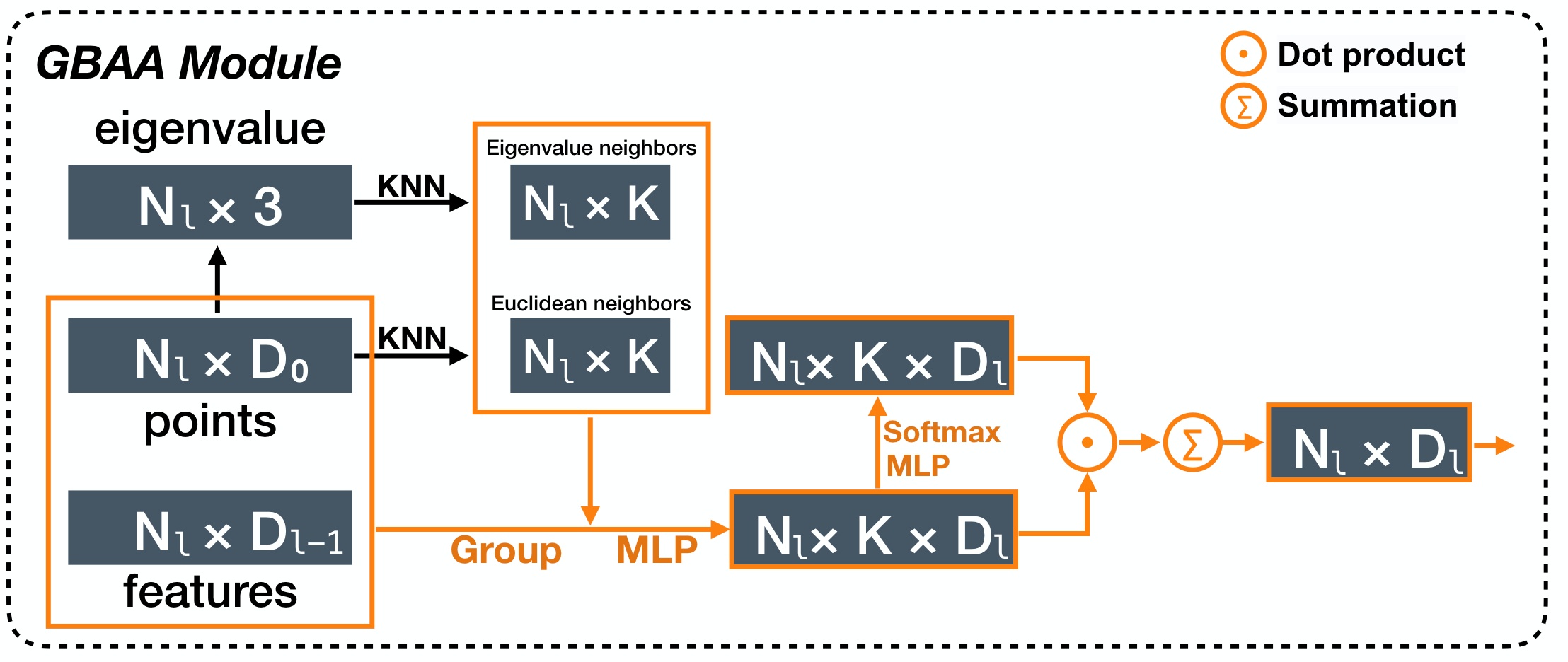} 
\caption{Encoder Module: Geometry Based Attentive Aggregation. The input are point-wise coordinates, colors and features. In GBAA, we aggregate features in both eigenvalue space and Euclidean space, then use attentive pooling to generate the output feature of each point.}
\label{fig_method_GSCA}
\end{figure}

This subsection describes the encoder module GBAA (Geometry Based Attentive Aggregation) of our network architecture. As Figure \ref{fig_method_GSCA} shows, the GBAA module is constituted of local feature aggregation and attentive pooling operations.

\noindent\textbf{Local Feature Aggregation}

In order to get a better description of each point, we enhance the local features with eigenvalues at each point. We utilize KNN algorithm to get the $K$-nearest neighbors of each point in Euclidean space. As introduced in \cite{Xu_2020}, we use the coordinates of neighbors of point $p_i$ to get eigenvalue-tuples denoted by $(\lambda_i^1,\lambda_i^2,\lambda_i^3)$. The original input features of each point  are denoted as $x^{0}_i=(\lambda_i^1,\lambda_i^2,\lambda_i^3)$.
For GBAA module in the $l$-th layer, we take original points $P$ and output of last layer $\mathbf{X}^{l-1}$ as input. We choose the $K$-nearest neighbors in Euclidean space and eigenvalue space respectively for each point $p_i$. Let $\{p_{i_{1}},...,p_{i_{K}}\}$ be the $K$-nearest neighbors of $p_i$ in Euclidean space, and their corresponding features are 
$\{x_{i_{1}}^{l-1},...,x_{i_{K}}^{l-1}\}$.
The features of $K$-nearest neighbors in eigenvalue space of point $p_i$ are 
$\{x_{\tilde{i_1}}^{l-1},...,x_{\tilde{i_{K}}}^{l-1}\}$. We define the local feature aggregation operation as $g^1_{\Theta_l}:\mathrm{R}^{2\times(3+d)}\times\mathrm{R}^{2\times D_{l-1}}\rightarrow\mathrm{R}^{D_{l}}$, where $g^1_{\Theta_l}$ is a nonlinear function with a set of learnable parameters $\Theta_l$, and $D_{l-1},D_{l}$ are dimension of output features of $l$-th and ($l$-$1$)-th layer respectively. In our module, $g^1_{\Theta_l}$ is a two-layer 2D-convolution. The local feature aggregation for each point is 
\begin{equation}
    x_{i}^{local,l}=g^1_{\Theta_l}(\|_{k=1}^{K}(p_{i_{k}}-p_{i})\oplus p_{i}\oplus x_{i_{k}}^{l-1}\oplus x_{\tilde{i_{k}}}^{l-1}).
\end{equation}
where $1\leq i \leq N$, $\oplus$ is concatenation and $x_{i}^{local,l}\in R^{K \times D_l}$. $\|$ is the concatenation among K dimension.

\noindent\textbf{Attentive Pooling}

For each point $p_i$, its local feature aggregation is $x^{local,l}_{i}\in \mathrm{R}^{K \times D_{l} }$. Instead of max pooling or average pooling, we apply an attentive pooling to $x_{i}^{local,l}$. 
\begin{equation}
    x_{i}^{l}=\sum\nolimits_{k=1}^{K}g^2_{\Theta_l}(x_{i}^{local,l}[k])\cdot x_{i}^{local,l}[k].
\end{equation}
where $g^2_{\Theta_l}$ is a 1-layer 2D convolution.

Moreover, as Figure \ref{fig_method_network} shows, we utilize the local feature aggregation and attentive pooling to obtain features of each point from two different receptive fields in order to enhance the representation of each point. The receptive field depends on $K$. We choose $K_1$ and $K_2$ nearest neighbors for feature aggregation and attentive pooling. Finally, we get the output of $l$-th layer which denoted as $\mathbf{X}^{l}=\mathbf{X}^{l}_{K_1}+\mathbf{X}^{l}_{K_2} $ .
\subsection{Decoder Module: FP \& IAF} \label{Sec_methods_FPIAF}
\begin{figure}[t]
\centering
\includegraphics[width=1\columnwidth]{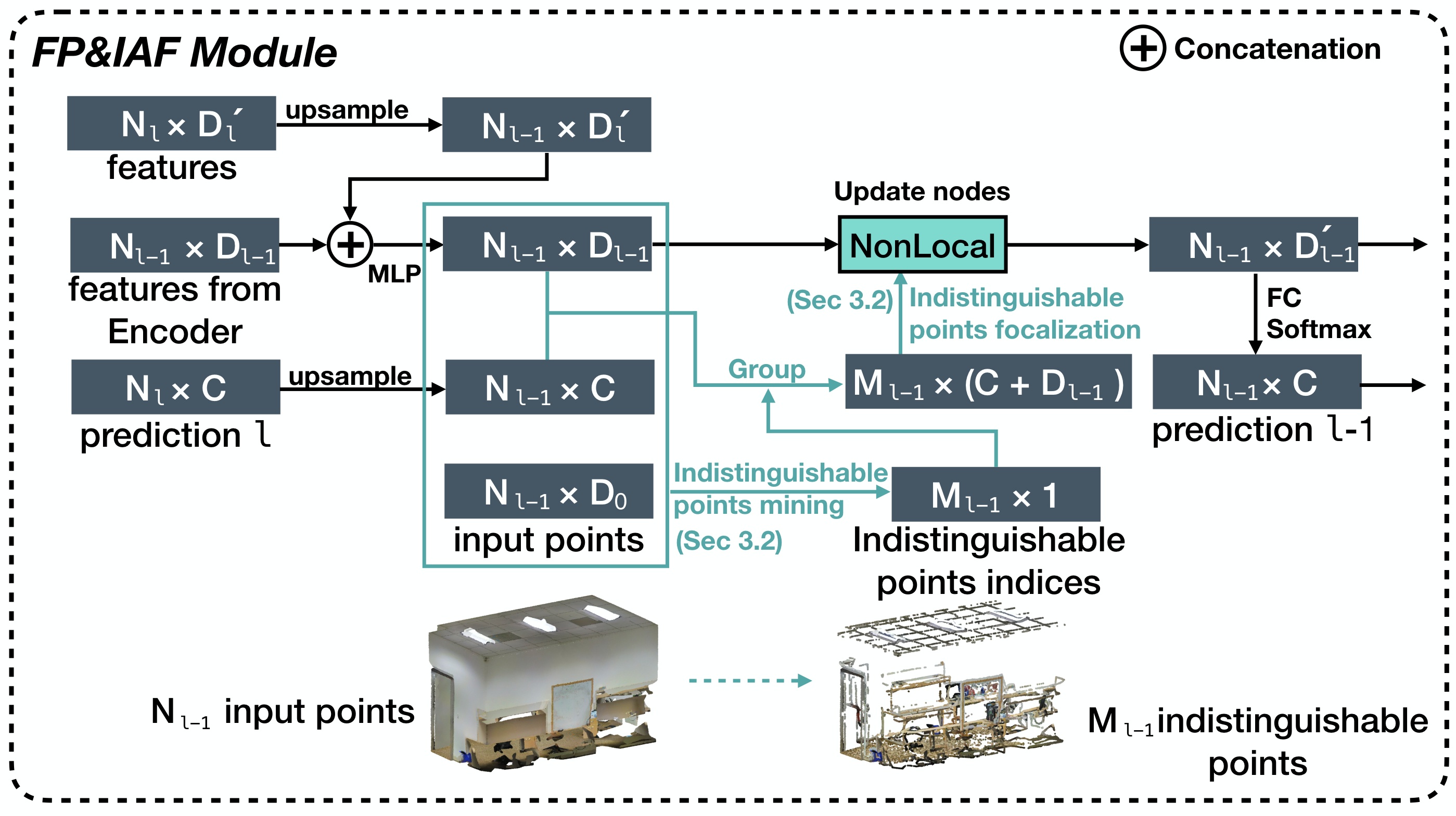} 
\caption{Decoder Module: It contains two stages: feature propagation and indistinguishable areas focalization.}
\label{fig_method_IAF}
\end{figure}

This subsection elucidates the decoder module FP \& IAF which is shown in Figure \ref{fig_method_IAF}. For convenience, we use ${{Y}^l}$ to represent the features of decoder module in the $l$-th layer.
The decoder module contains two parts: feature propagation and indistinguishable areas focalization (IAF).

\noindent\textbf{Feature Propagation} 

In encoder part, the original point set is sub-sampled. We adopt a hierarchical propagation strategy with distance based interpolation and across level skip links as \cite{qi2017pointnet++}. In a feature propagation process, we propagate point features $Y^{l}\in \mathrm{R}^{N_{l}\times D_{l}'}$ and label predictions $Z^{l}\in \mathrm{R}^{N_{l}\times C}$ to $Y^{l-1}_{up}\in \mathrm{R}^{N_{l-1}\times D_{l-1}'}$ and $Z^{l-1}_{up}\in \mathrm{R}^{N_{l-1}\times C}$, where $N_{l}$ and $N_{l-1}$ ( $N_{l}\leq N_{l-1}$) are point set size of input and output of the $l$-th layer, $C$ is the number of categories for semantic segmentation.
\begin{equation}
    y^{l-1}_{i,fp}=g_{\Psi_{l-1}}(y^{l}_{i,up} \oplus x^{l-1}_{i}).
\end{equation}
where $g_{\Psi_{l-1}}$ is convolution operations with batch normalization and activate-function, $x_{i}^{l-1}$ is the features from the encoder module in the ($l$-$1$)-th layer.

\noindent\textbf{Indistinguishable Areas Focalization}

\textbf{Indistinguishable points mining.}
In order to discover indistinguishable points adaptively in the training process, both low level geometric and high level semantic information can be used to mine these points. 
Local difference is the difference between each point and its neighbors. To a certain extent, local difference reflects the distinctiveness of each point which depend on low-level geometry, latent space and high-level semantic features. So we use local difference as a criterion of mining indistinguishable points.
For each point $p_{i}$, we get the $K$-nearest neighbors in Euclidean space, then we have the following local difference of each point in each layer:
\begin{equation}
    LD_1^l(p_{i})=\sum\nolimits^{K}_{k=1}||p_{i}-p_{i_{k}}||_{2}.
\end{equation}
\begin{equation}
    LD_{2}^l(p_{i})=\sum\nolimits_{k=1}^{K}||z_{i,up}^{l-1}-z_{i_{k},up}^{l-1}||_{2}.
\end{equation}
\begin{equation}
    LD_{3}^l(p_{i})=\sum\nolimits_{k=1}^{K}||y_{i,fp}^{l-1}-y_{i_{k},fp}^{l-1}||_{2}.
\end{equation}
Then we accumulate these local differences together:
\begin{equation}
    LD^l(p_i)= \sum\limits_{j=1}^{3} \mu_{j} \times \frac{LD^l_j(p_i)-\min(LD^l_j(p))}{\max(LD^l_j(p))-\min(LD^l_j(p))}.
\end{equation}
where $0\leq\mu_j\leq1$.

\begin{figure}[t]
\centering
\includegraphics[width=1\columnwidth]{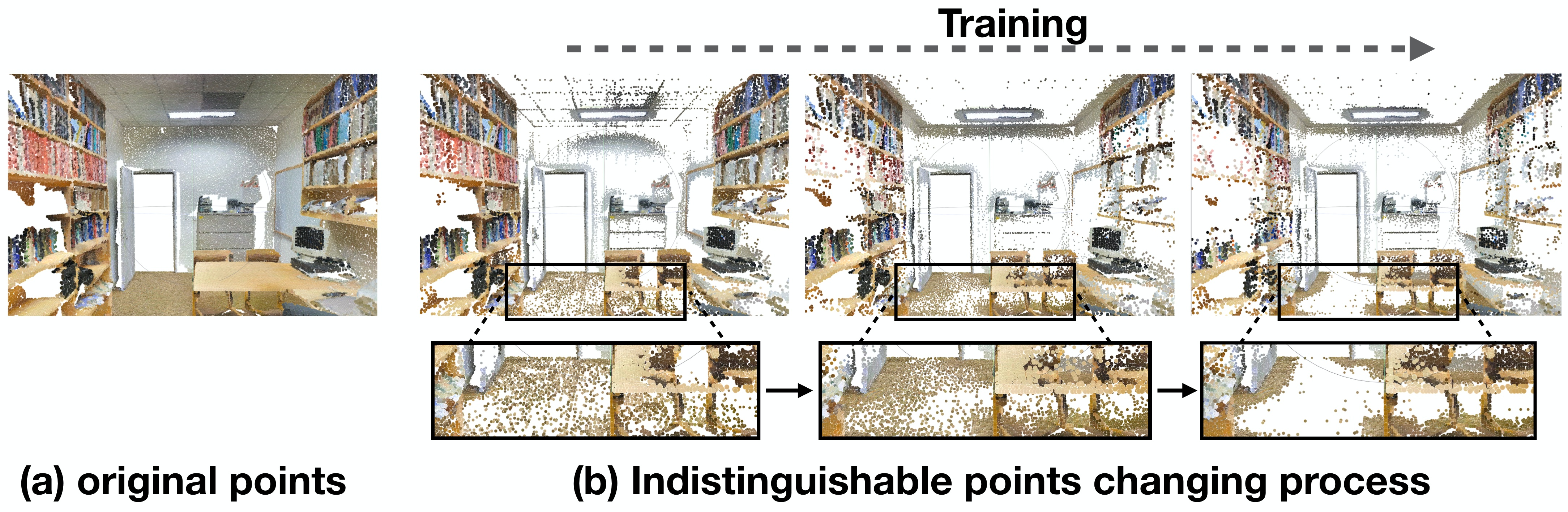} 
\caption{The adaptive change process of indistinguishable points during the training process. The background is colored white.  \textbf{Best viewed in color with 300\% zoom.}}
\label{fig_vis_adaptive}
\end{figure}

$LD^l$ indicates the accumulation of fine-grained features' difference $LD^l_{3}$ among each point's local region, high-level semantic predictions' local difference $LD^l_{2}$ and low-level properties' local difference $LD^l_{1}$, where $\{\mu_j\}$ is used to adjust the weight of these three local differences.
We align the points in a descending order according to $LD^l$, then choose top $M_{l-1}=\frac{N_{l-1}}{\tau}$ points as the indistinguishable points.
There are three types of points mentioned in Introduction as Figure \ref{fig_introduction_2}, \ref{fig_method_IAF} shows.
These indistinguishable points change dynamically as the network updates iteratively (Figure \ref{fig_vis_adaptive}).
It is noted that at the beginning of training, the indistinguishable points are distributed over the areas where the original properties (coordinates and colors) change rapidly.
As the training process goes on, the indistinguishable points locate at the indistinguishable areas mentioned in the introduction.


\textbf{Indistinguishable points set focalization.}
We aggregate intermediate features and label predictions of the indistinguishable points, then use the MLP \cite{hornik1991approximation} to extract the features for indistinguishable points separately.
\begin{equation}
    x^{l-1}_{j\in M_{l-1}}=g^1_{\Omega_{l}}(y^{l-1}_{j,fp}\oplus z^{l-1}_{j,up})\in \mathrm{R}^{D_{l-1}}.
\end{equation}
where $j\in M_{l-1}$ means that the points belong to the indistinguishable points set and $g^1_{\Omega_{l}}$ is MLP operations. 

\textbf{Update nodes.}
To enhance the features of points, especially the indistinguishable points,
here we utilize the NonLocal mechanism to update the features of all points by the following equations and it can enhance the features of indistinguishable points implicitly.
\begin{equation}
    y^{l-1}_{i}=g^2_{\Omega_{l}}(\sum\limits_{j\in M_{l-1}}(g^3_{\Omega_{l}}(x^{l-1}_{j})\odot g^4_{\Omega_{l}}(y^{l-1}_{i,fp}))\cdot g^5_{\Omega_{l}}(y^{l-1}_{i,fp})).
\end{equation}
where $g^2_{\Omega_{l}}, g^3_{\Omega_{l}}, g^4_{\Omega_{l}}, g^5_{\Omega_{l}}$ are MLPs.
Also, we have the label prediction probability $z^{l-1}_{i}$ of point $p_{i}$ in ($l$-1)-th layer. 
\begin{equation}
    z^{l-1}_{i}=Softmax(g^{6}_{\Omega_{l-1}}(y_{i}^{l-1})\in \mathrm{R}^{C}).
\end{equation}

\subsection{Loss for Segmentation}

In order to progressively refine the features of indistinguishable areas, we apply the multi-stage loss as follows,
\begin{equation}
    L_{ms}^{l}=CrossEntropy(Z_{gt}^{l},Z^{l}).
\end{equation}
where $Z_{gt}^{l} \in \mathrm{R}^{N^l \times 1} $ is the ground truth points' labels in $l$-th layer.

As the output of the last layer is $y^{1}_{i}$, inspired by \cite{Han_2020}, we use the self correlation, local correlation and nonlocal correlation operation to augment features of each point $p_{i}$. Finally, we get features of each point $p_{i}$ as the accumulation of three correlations' outputs.
The final loss for training is as follows:
\begin{equation}
    L_{f} = \sum\limits_{l=1}^{5}L_{ms}^{l} + L_{p}.
\end{equation}
where the $L_{p}$ is the loss between final label predictions $Z$ and ground truth labels $Z_{gt}$.


\begin{figure}[t]
    \centering
    \includegraphics[width=1\columnwidth]{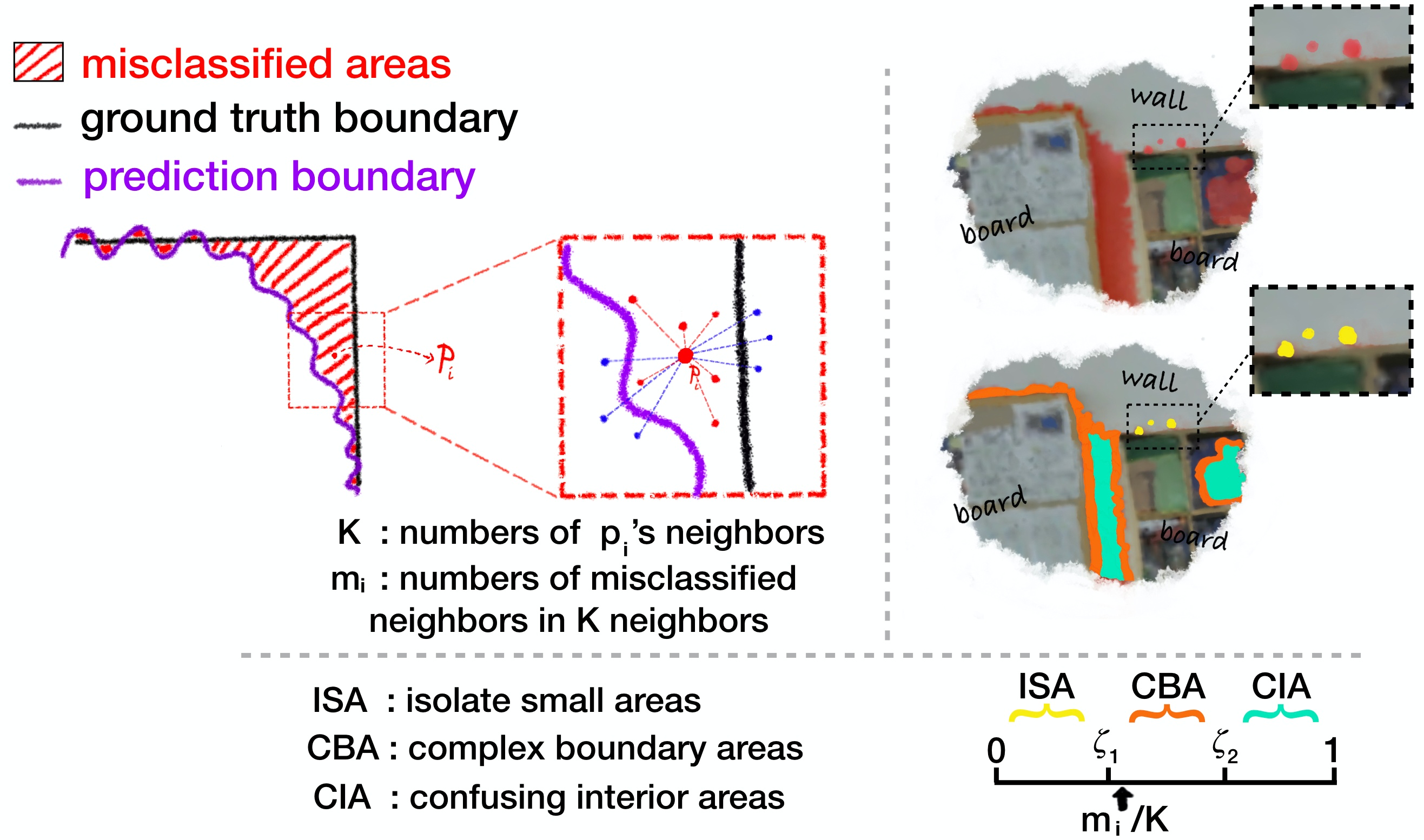} 
    \caption{The evaluation process of the indistinguishable points based metric (IPBM). The black line is the boundary of the ground truth, the purple is the prediction boundary, and the red areas are misclassified areas. $\zeta_1, \zeta_2 $ are the parameters for partitioning the three types of indistinguishable points.}
    \label{fig_method_IBM}
\end{figure}

\subsection{Indistinguishable Points Based Metric} \label{Sec_methods_IPBM}
To better distinguish the effect of different methods in 3D semantic segmentation, we propose a novel evaluation metric named ``Indistinguishable Points Based Metric'' (IPBM). This evaluation metric focuses on the effectiveness of segmentation methods on the indistinguishable areas.

For the whole points $\mathrm{P}=\{p_1,p_2,...,p_N\}$, we have the predictions $Pred = \{z_{i}, 1\leq i\leq N \}$ and ground truth labels $Label = \{z_{i,gt}, 1\leq i\leq N \}$ .
Figure \ref{fig_method_IBM} shows the processing details of the IPBM.
Firstly, for point $p_i$ satisfying the factor $z_{i} \neq z_{i,gt}$, its neighbors in Euclidean space are $\{z_{i_j}, 1\leq j\leq K\}$. Then we denote the number of neighbor points that satisfy $z_{i_k} \neq z_{i_k,gt}$ as $m_i$ for each point $p_i$.
Next, we divide interval [0, 1] (domain of $\frac{m_i}{K}$) into three partitions with endpoints as $0, \zeta_{1}, \zeta_{2}, 1$.

\begin{figure}[t]
    \centering
    \includegraphics[width=0.66\columnwidth]{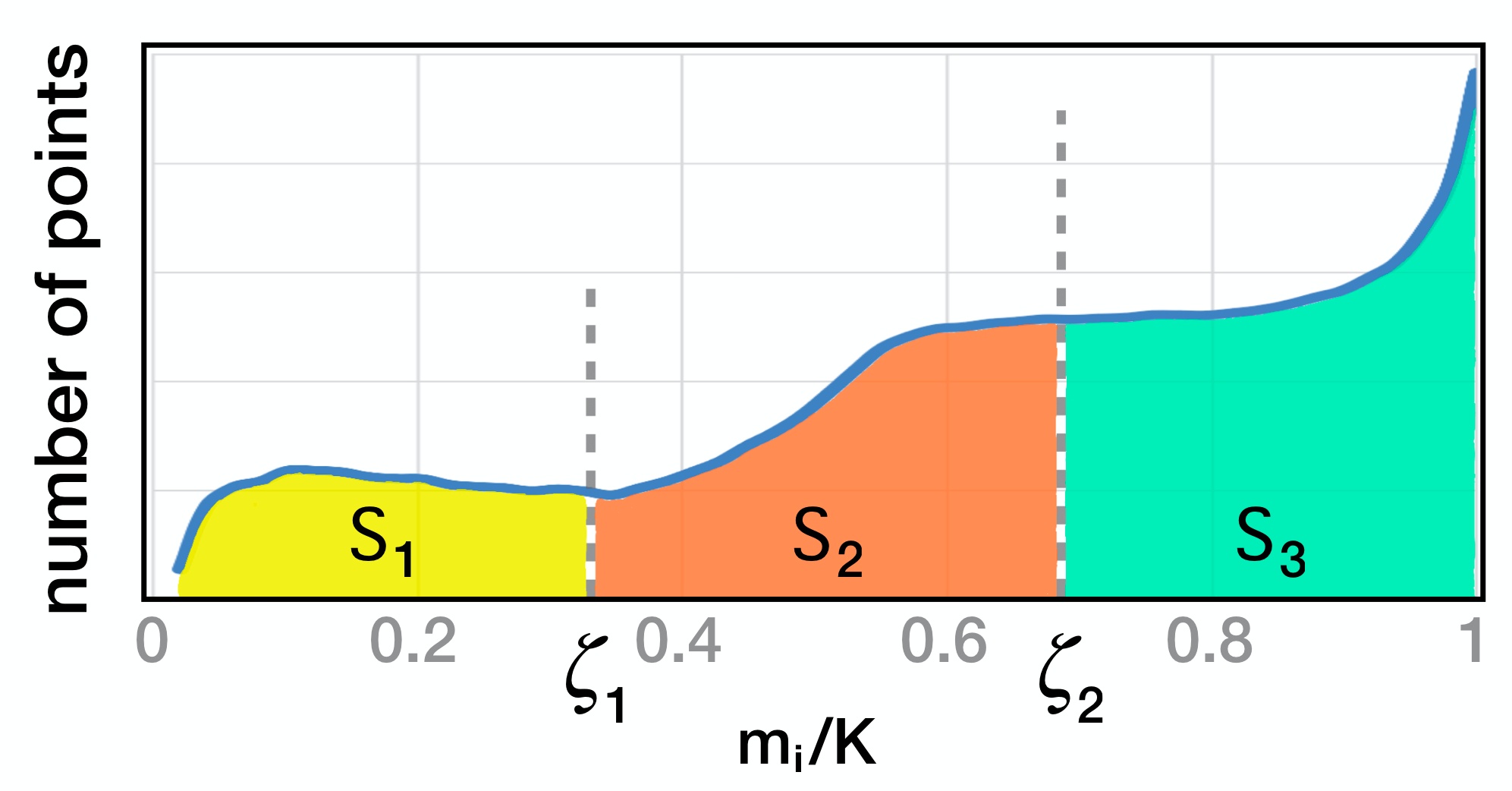} 
    \caption{The curve of the points number as $\frac{m_{i}}{K}$ changes on S3DIS dataset. $\zeta_1=0.33, \zeta_2=0.66$. $S_1,S_2,S_3$ are number of points in three types of indistinguishable areas. }
    \label{fig_method_IPBM_Partition_line}
\end{figure}
\begin{figure}[t]
    \centering
    \includegraphics[width=1\columnwidth]{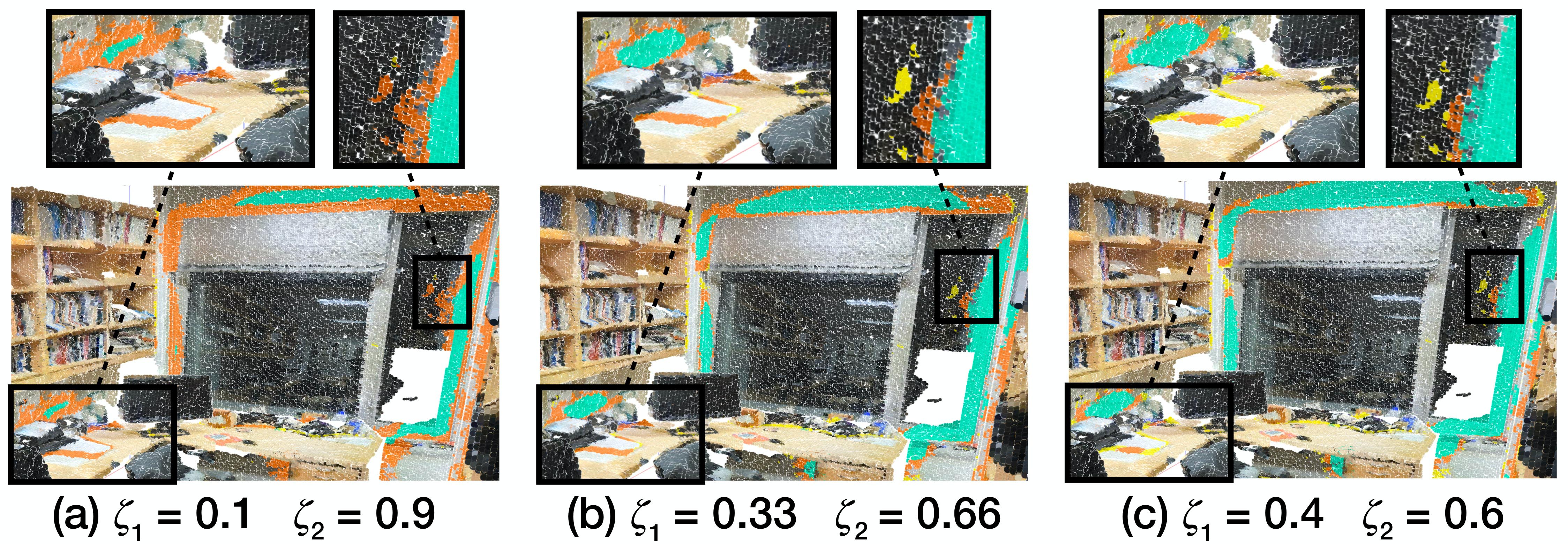} 
    \caption{The visual experiments of $\zeta_1, \zeta_2$ on S3DIS dataset. Yellow areas are isolate small areas; Orange areas are complex boundary areas; Cyan areas are confusing interior areas. \textbf{Best viewed in color with 300\% zoom.}}
    \label{fig_method_IPBM_Partition}
\end{figure}

We determine $\zeta_1=0.33, \zeta_2=0.66 $ empirically by considering the curve in Figure \ref{fig_method_IPBM_Partition_line}. To be more specific, Figure \ref{fig_method_IPBM_Partition_line} shows that the growth trend of number of points with value $\frac{m_{i}}{K}$ on S3DIS dataset. The curve can be divided into three partitions. From a large number of visualizations, we find that these three partitions roughly reflect the value distribution of the three types of indistinguishable areas. The examples of visualization under different choices of $\zeta_1, \zeta_2 $ are shown in Figure \ref{fig_method_IPBM_Partition}.

Finally, we use $\frac{S_1}{N}$, $\frac{S_2}{N}$, $\frac{S_3}{N}$ as our new evaluation metric where $S_1,S_2,S_3$ are number of points in three types of indistinguishable areas. As Figure \ref{fig_method_IPBM_Partition} shows, $\frac{S_1}{N}$ is used to evaluate the method's performance on isolate small areas (colored yellow), $\frac{S_2}{N}$ is for complex boundary areas (colored orange), and $\frac{S_3}{N}$ is for confusing interior areas (colored cyan).

\begin{figure}[t]
    \centering
    \includegraphics[width=1\columnwidth]{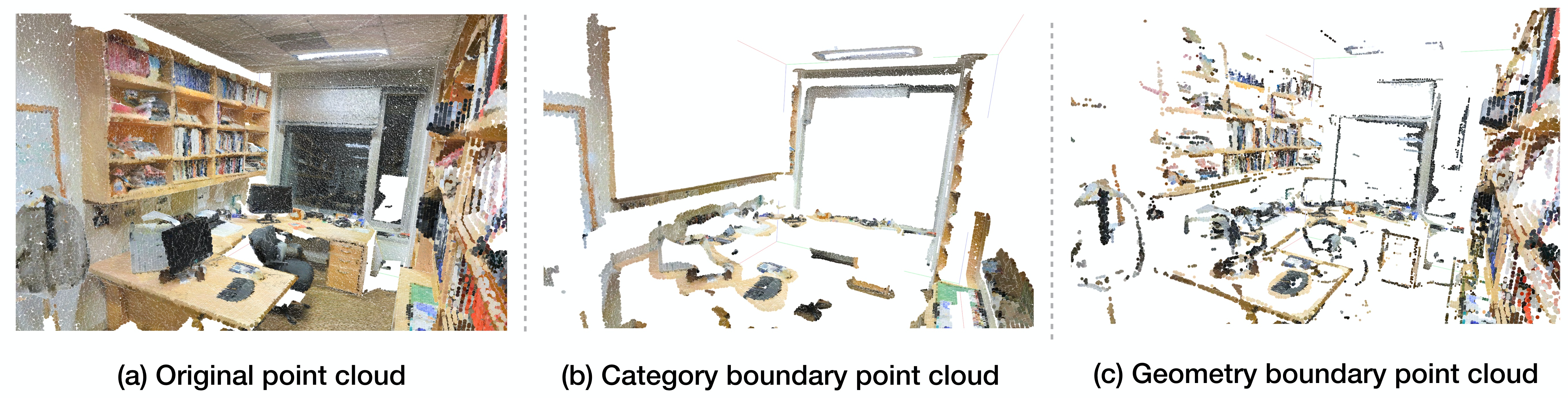} 
    \caption{Three subsets of the point cloud used for the evaluation on indistinguishable points based metric. The background is colored white. }
    \label{fig_method_IBM_3pts}
\end{figure}

For a more comprehensive evaluation, three subsets of the point cloud are sampled for the above evaluation. As Figure \ref{fig_method_IBM_3pts} shows, they are original point cloud, category boundary point cloud and geometry boundary point cloud. The specific methods of subset point cloud acquisition is explained in the supplementary materials.


\begin{table*}[]
    \resizebox{\textwidth}{!}{
    \begin{tabular}{l|ccc|ccccccccccccc}
        \hline \text {Methods (published time order)} & \text {OA (\%)} & \text {mAcc (\%)} & \text {mIoU (\%)} & \text {ceiling} & \text {flooring} & \text {wall} & \text {beam} & \text {column} & \text {window} & \text {door} & \text {table} & \text {chair} & \text {sofa} & \text {bookcase} & \text {board} & \text {clutter} \\
        \hline
        \text {PointNet \cite{qi2017pointnet}} & - & 49.0 & 41.1 & 88.8 & 97.3 & 69.8 & 0.1 & 3.9 & 46.3 & 10.8 & 58.9 & 52.6 & 5.9 & 40.3 & 26.4 & 33.2\\         
        \text{SegCloud \cite{tchapmi2017segcloud} } & - & 57.4 & 48.9 & 90.1 & 96.1 & 69.9 & 0.0 & 18.4 & 38.4 & 23.1 & 70.4 & 75.9 & 40.9 & 58.4 & 13.0 & 41.6\\
        \text{TangentConv \cite{tatarchenko2018tangent}} & - & 62.2 & 52.6 & 90.5 & 97.7 & 74.0 & 0.0 & 20.7 & 39.0 & 31.3 & 77.5 & 69.4 & 57.3 & 38.5 & 48.8 & 39.8\\
        \text {SPGraph \cite{landrieu2018large}} & 86.4 & 66.5 & 58.0 & 89.4 & 96.9 & 78.1 & 0.0 & \textbf{42.8} & 48.9 & 61.6 & \underline{84.7} & 75.4 & 69.8 & 52.6 & 2.1 & 52.2 \\
        \text{PCNN \cite{wang2018deep}} & - & 67.1 & 58.3 & 92.3 & 96.2 & 75.9 & {0.3} & 6.0 & \textbf{{69.5}} & \underline{63.5} & 65.6 & 66.9 & 68.9 & 47.3 & 59.1 & 46.2\\        
        \text{RNNFusion \cite{ye20183d}} & - & 63.9 & 57.3 & 92.3 & 98.2 & 79.4 & 0.0 & 17.6 & 22.8 & 62.1 & 80.6 & 74.4 & 66.7  & 31.7& 62.1 & \underline{56.7}\\
        \text{Eff 3D Conv \cite{zhang2018efficient}} & - & 68.3 & 51.8 & 79.8 & 93.9 & 69.0 & 0.2 & 28.3 & 38.5 & 48.3 & 73.6 & 71.1 & 59.2 & 48.7 & 29.3 & 33.1 \\
        \text {PointCNN \cite{li2018pointcnn}} & 85.9 & 63.9 & 57.3 & 92.3 & 98.2 & 79.4 & 0.0 & 17.6 & 22.8 & 62.1 & 74.4 & 80.6 & 31.7 & 66.7 & 62.1 & \underline{56.7} \\
        \text {PointWeb \cite{zhao2019pointweb}}  & 87.0 & 66.6 & 60.3 & 92.0 & \underline{98.5} & 79.4 & 0.0 & 21.1 & 59.7 & 34.8 & 76.3 & \underline{88.3} & 46.9 & 69.3 & 64.9 & 52.5 \\
        \text{GACNet \cite{wang2019graph}} & 87.8 & - & 62.9 & 92.3 & 98.3 & 81.9 & 0.0 & 20.4 & 59.1 & 40.9 & \textbf{{85.8}} & 78.5 & \underline{70.8} & 61.7 & \underline{74.7} & 52.8 \\
        \text{KPConv \cite{thomas2019kpconv}} & - & \textbf{72.8} & \textbf{67.1} & {92.8} & 97.3 & \underline{82.4} & 0.0 & 23.9 & 58.0 &\textbf{ 69.0} & 81.5 & \textbf{91.0} & \textbf{75.4} & \textbf{75.3} & 66.7 & \textbf{58.9} \\        
        \text {Point2Node \cite{Han_2020}}  & \textbf{88.8} & 70.0 & 63.0 & \underline{93.9} & 98.3 & \textbf{83.3} & 0.0 & \underline{35.7} & 55.3 & 58.8 & 79.5 & 84.7 & 44.1 & 71.1 & 58.7 & 55.2 \\     
        \text{FPConv \cite{lin2020fpconv}} & - & - & 62.8 & \textbf{94.6} & \underline{98.5} & 80.9 & 0.0 & 19.1 & 60.1 & 48.9 & 80.6 & 88.0 & 53.2 & 68.4 & 68.2 & 54.9\\
        \hline
        \textbf{Ours(IAF-Net)} & \underline{88.4} & \underline{70.4} & \underline{64.6}& 91.4 & \textbf{98.6} & 81.8 & 0.0 & 34.9 & \underline{62.0} & 54.7 & 79.7 & 86.9 & 49.9 & \underline{72.4} & \textbf{74.8} & 52.1 \\
        \hline
    \end{tabular}
    }
    \caption{Semantic segmentation results on S3DIS dataset evaluated on Area 5.}
    \label{S3DIS_Area5}
\end{table*}

\section{Experiments}
\subsection{Experimental Evaluations on Benchmarks}
\textbf{S3DIS Semantic Segmentation}

\textbf{Dateset.}
The S3DIS \cite{armeni20163d} dataset contains six sets of point cloud data from three different buildings (including 271 rooms). 
We follow \cite{boulch2020convpoint} to prepare the dataset. 

For training, we randomly select points in the considered point cloud, and extract all points in an infinite column centered on this point, where the column section is 2 meters. For each column, we randomly select 8192 points as the input points. During the testing, for a more systematic sampling of the space, we compute a 2D occupancy pixel map with pixel size 0.1 meters. Then, we consider each occupied cell as a center for a column (same for training). Finally, the output scores are aggregated (sum) at point level and points not seen by the network receive the label of its nearest neighbors. Following \cite{boulch2020convpoint}, we report the results under two settings: testing on Area 5 and 6-fold cross validation. 

\textbf{Performance Comparison.}
Table \ref{S3DIS_Area5} and Table \ref{S3DIS_6fold} show the quantitative results of different methods under two settings mentioned above, respectively. 
Our method achieves on-par performance with the SOTA methods. It is noted that some methods \cite{qi2017pointnet, zhao2019pointweb} use small column section(1 meter), and it may not contain enough holistic information. It will be more common to the situation which the column section cannot contain the whole object.
By contrast, our method use IAF module to deal with the indistinguishable points specially and use big column section (2 meters) for getting more geometry information as the input. 
For the Area 5 evaluation, our method achieves the best performance except KPConv \cite{thomas2019kpconv}, and get 2.01{\%} higher result than Point2Node \cite{Han_2020}. For the 6-fold evaluation, our method achieves comparable performance (70.3{\%}) with the state-of-the-art method \cite{thomas2019kpconv}.The parameters of KPConv is 14.9M, while our IAF-Net use less parameters (10.98M).
Besides, we do not use voting test due to the large scale points in S3DIS, it takes a huge amount of computing resources and time.

\noindent\textbf{ScanNet Semantic Voxel Labeling}

The ScanNet \cite{dai2017scannet} dataset contains 1,513 scanned and reconstructed indoor scenes, split into 1,201/312 for training/testing. For the semantic voxel labeling task, 20 categories are used for evaluation and 1 class for free space. We followed the same data pre-processing strategies as with \cite{zhao2019pointweb}, where points are uniformly sampled from scenes and are divided into blocks, each of size 1.5m×1.5m. During the training, 8,192 point samples are chosen, where no less than 2\% voxels are occupied and at least 70\% of the surface voxels have valid annotation. Points are sampled on-the-fly. All points in the testing set are used for evaluation and a smaller sampling stride of 0.5 between each pair of adjacent blocks is adopted during the testing. In the evaluation, overall semantic voxel labeling accuracy is adopted. For fair comparisons with the previous approaches, we do not use the RGB color information for training and testing. Table \ref{ScanNet_result} shows the semantic voxel labeling results. Our method achieves comparable performance (85.8{\%}) with the state-of-the-art methods on ScanNet dataset.

\begin{table}
	\centering
	\scriptsize
	\begin{tabular}{c|c|ccc}
		\hline
		Subsets &Methods & ISA (\%)& CBA (\%) & CIA (\%)\\
		\hline
		&PointWeb & 1.48 & 2.83 & 9.38 \\
        original &KPConv & 1.33 & 2.46  & 9.28 \\
         point cloud & RandLANet & 1.23 & 2.58 & 9.07  \\
		&\textbf{Ours(IAF-Net)} & \textbf{1.08} & \textbf{2.03} & \textbf{8.46}\\
		\hline\hline
		&PointWeb  & 3.98 & 6.94 & 14.31 \\
        category&KPConv & 2.73 & 4.71 & 14.21 \\
         boundary & RandLANet & 2.57 & 5.19 & 15.02  \\
		&\textbf{Ours(IAF-Net)} & \textbf{2.40} & \textbf{4.33} &\textbf{ 13.76}\\
		\hline\hline
		&PointWeb & 2.75 & 4.47 & 12.94 \\
        geometry&KPConv & 4.60 & 6.13 & \textbf{10.89} \\
        boundary & RandLANet & 2.32 & 4.02 & 13.23  \\
		&\textbf{Ours(IAF-Net)} & \textbf{2.06} & \textbf{3.37} & 12.42\\	
		\hline
	\end{tabular}
	\caption{Results on Indistinguishable Points Based Metric (IPBM). 'ISA': isolate small areas; 'CBA': complex boundary areas; 'CIA': confusing interior areas.}
	\label{IPBM_result}
\end{table}

\begin{table*}[]
    \resizebox{\textwidth}{!}{
    \begin{tabular}{l|ccc|ccccccccccccc} 
        \hline \text {Methods (published time order)} & \text {OA (\%)} & \text {mAcc (\%)} & \text {mIoU (\%)} & \text {ceiling} & \text {flooring} & \text {wall} & \text {beam} & \text {column} & \text {window} & \text {door} & \text {table} & \text {chair} & \text {sofa} & \text {bookcase} & \text {board} & \text {clutter} \\
        \hline \text {PointNet \cite{qi2017pointnet}} & 78.5 & 66.2 & 47.8 & 88.0 & 88.7 & 69.3 & 42.4 & 23.1 & 47.5 & 51.6 & 54.1 & 42.0 & 9.6 & 38.2 & 29.4 & 35.2 \\
        \text {DGCNN\cite{wang2018dynamic}} & 84.1 & - & 56.1 & - & - & - & - & - & - & - & - & - & - & - & - & - \\        
        \text {RSNet \cite{huang2018recurrent}} & - & 66.5 & 56.5 & 92.5 & 92.8 & 78.6 & 32.8 & 34.4 & 51.6 & 68.1 & 59.7 & 60.1 & 16.4 & 50.2 & 44.9 & 52.0 \\
        \text{PCNN \cite{wang2018deep}} & - & 67.0 & 58.3 & 92.3 & 96.2 & 75.9 & 0.27 & 6.0 & \underline{69.5} & 63.5 & 66.9 & 65.6 & 47.3 & \underline{68.9} &59.1 & 46.2\\ 
        \text {SPGraph \cite{landrieu2018large}} & 85.5 & 73.0 & 62.1 & 89.9 & 95.1 & 76.4 & 62.8 & 47.1 & 55.3 & 68.4 & 69.2 & 73.5 & 45.9 & 63.2 & 8.7 & 52.9 \\
        \text {PointCNN \cite{li2018pointcnn}} & 88.1 & 75.6 & 65.4 & \textbf{94.8} & {97.3} & 75.8 & \underline{63.3} & 51.7 & 58.4 & 57.2 & 69.1 & 71.6 & 61.2 & 39.1 & 52.2 & 58.6 \\   
        \text{A-CNN \cite{Komarichev_2019}} & 87.3 & - & 62.9 & 92.4 & 96.4 & 79.2 & 59.5 & 34.2 & 56.3 &65.0 & 66.5 & {78.0} & 28.5 &56.9 & 48.0 & 56.8 \\
        \text {PointWeb \cite{zhao2019pointweb}} & 87.3 & 76.2 & 66.7 & 93.5 & 94.2 & 80.8 & 52.4 & 41.3 & 64.9 & 68.1 & 71.4 & 67.1 & 50.3 & 62.7 & 6.2 & 58.5 \\
        \text{KPConv \cite{thomas2019kpconv}} & - & \underline{79.1} & \textbf{70.6} & 93.6 & 92.4 & \underline{83.1} & \textbf{63.9} & \textbf{54.3} & 66.1 & \textbf{76.6} & 64.0 & 57.8 & \textbf{74.9} & \textbf{69.3}  & 61.3 & 60.3\\  
        \text{ShellNet \cite{zhang-shellnet-iccv19}} &  87.1 & - & 66.8 & 90.2 & 93.6 & 79.9 & 60.4 & 44.1 & 64.9 & 52.9 & 71.6 & \textbf{84.7} & 53.8 & 64.6 & 48.6 & 59.4 \\
        \text {Point2Node \cite{Han_2020}} & \textbf{{89.0}} & \underline{{79.1}}& {70.0} & \underline{94.1} & {97.3} & \textbf{{83.4}} & 62.7 & \underline{52.3} & \textbf{{72.3}} & 64.3 & \textbf{{75.8}} & 70.8 & \underline{65.7} & 49.8 & 60.3 & \underline{{60.9}} \\    
        \text{RandLA-Net \cite{Hu_2020}} & 87.1 & \textbf{{81.5}} & 68.5 & 92.7 & 95.6 & 79.2 & 61.7 & 47.0 & 63.1 & 67.7 & 68.9 & 74.2 & 55.3 & 63.4 & {63.0} & 58.7\\
        \text{FPConv \cite{lin2020fpconv}} & - & - & 68.7 & \textbf{94.8} & \underline{97.5} & 82.6 & 42.8 & 41.8 & 58.6 & 73.4 & 71.0 & \underline{81.0} & 59.8 & 61.9 & \underline{64.2} & \textbf{64.2}\\
        \hline
        \textbf{Ours(IAF-Net)} & \underline{88.8} & 77.8 & \underline{70.3} & 93.3 & \textbf{97.9} & 81.9 & 55.2 & 42.7 & 64.9 & \underline{74.7} & \underline{74.2} & 71.8 & 63.3 & 66.2 & \textbf{66.5} & {60.5}\\
        \hline
    \end{tabular}
    }
    \caption{Semantic segmentation results on S3DIS dataset with 6-folds cross validation. }
    \label{S3DIS_6fold}
\end{table*}

\begin{table}
\small
	\centering
	\begin{tabular}{l|c}
		\hline
		Methods & OA (\%)\\
		\hline
		3DCNN \cite{bruna2013spectral} & 73.0\\
		PointNet \cite{Charles_2017} & 73.9 \\
		TCDP \cite{tatarchenko2018tangent} & 80.9 \\
		PointNet++ \cite{qi2017pointnet++} & 84.5\\
		PointCNN \cite{li2018pointcnn} & 85.1\\
		A-CNN \cite{Komarichev_2019} & 85.4 \\
		PointWeb \cite{zhao2019pointweb} & 85.9\\
		\hline
		\textbf{Ours(IAF-Net)} & \textbf{85.8}\\
		\hline
	\end{tabular}
	\caption{Results on ScanNet dataset.}
	\label{ScanNet_result}
\end{table}

\subsection{The Evaluation Results of IPBM}
As we have described in Sec. \ref{Sec_methods_IPBM}, we propose a novel evaluation metric (IPBM) for distinguishing the effect of different methods. We compare our method with the state-of-art methods on S3DIS dataset (Area 5 evaluation), and we use the prediction of the methods to generate the results under the IPBM. The results are summarized in Table \ref{IPBM_result} with three settings: original point cloud, category boundary and geometry boundary which correspond to three subsets of Sec 3.3 (shown in Figure \ref{fig_method_IBM_3pts}) respectively. All methods in Table \ref{IPBM_result} are reproduced by ourselves.
Our method achieves the best performance under the settings of original point cloud and geometry boundary, and get comparable result with KPConv \cite{thomas2019kpconv} under the setting of category boundary. More visualization results can be found in supplementary material.


\section{Analysis}
\subsection{Analysis on Indistinguishable Points Mining}
In this section, we conduct a series of experiments on the hyperparameters in the indistinguishable points mining process. 
As Section \ref{Sec_methods_FPIAF} shows, 
$LD^l$ is the weighted sum of three local differences, where the weight factors is $\{ \mu_1, \mu_2, \mu_3 \}$,
then we choose top $\frac{N_{l-1}}{\tau}$ points according to $LD^l$ as the indistinguishable points. 
The following experiments are tested on S3DIS dataset (Area 5 evaluation). 
The accumulation of local differences can be found in supplementary materials, and the proportion of the indistinguishable points in original points is introduced as follows.

\textbf{ Proportion of the indistinguishable points in original points.}
In order to achieve the balance of indistinguishable points and original points.
$\tau$ is used to control the proportion of indistinguishable points in input points in each layer. As Figure \ref{fig_ablation_numbers} shows, when we set the proportion as 1:4, we get the best performance. 
When the proportion is too large, it will increase training difficulty of NonLocal mechanism and then degrade the performance.
By contrast, when the proportion is too small, the indistinguishable points set may not cover all categories, because the indistinguishable points in different category may differ in degree.

\begin{figure}[t]
\centering
\includegraphics[width=1\columnwidth]{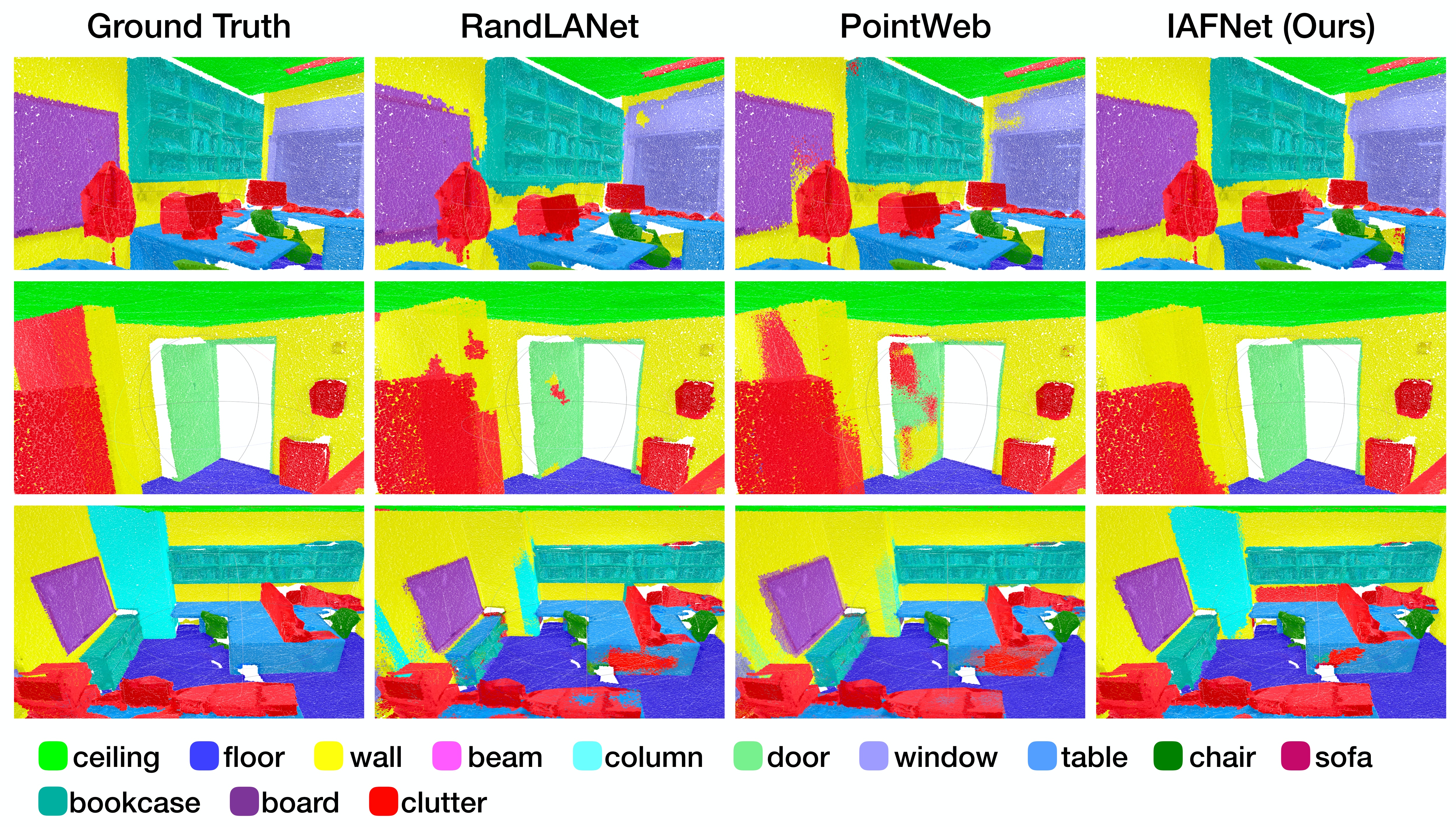} 
\caption{Visual comparison of semantic segmentation results on S3DIS dataset. \textbf{Best viewed in color and zoom in.}}
\label{fig_vis_result}
\end{figure}

\begin{figure}[t]
\centering
\includegraphics[width=0.7\columnwidth]{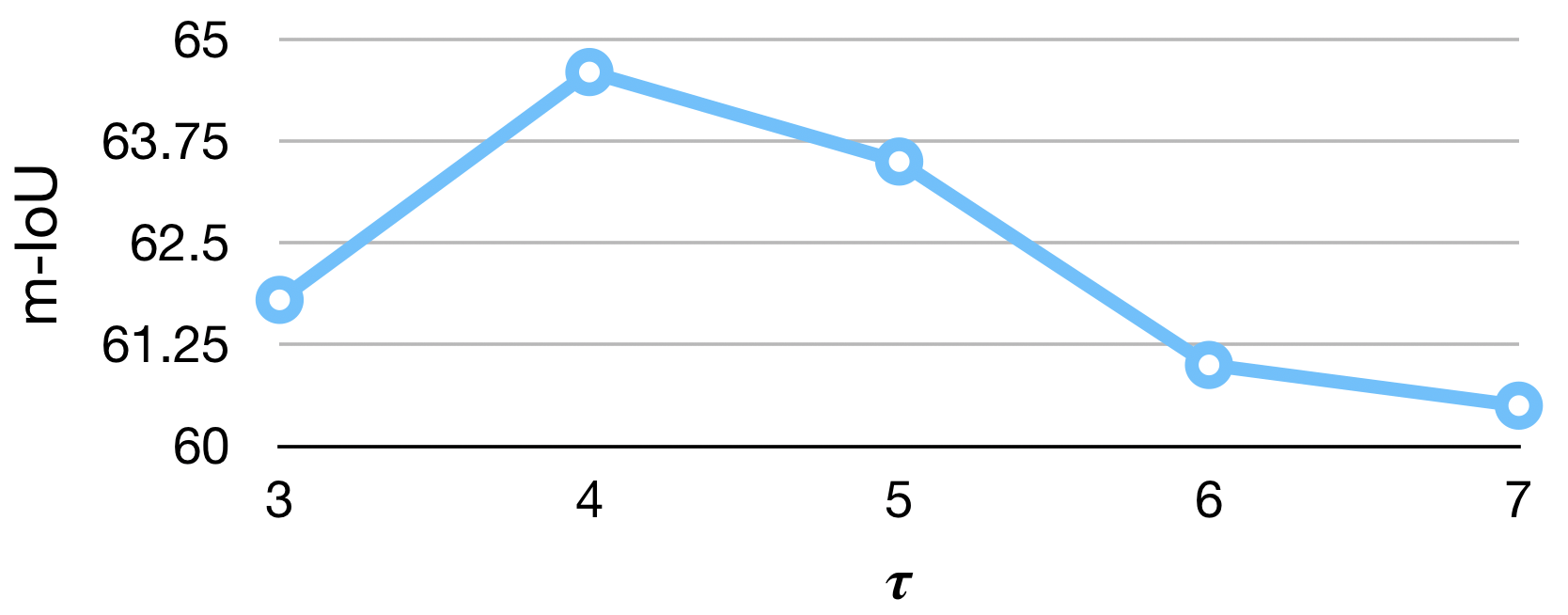} 
\caption{Evaluation of the m-IoU when reducing the number of indistinguishable points. }
\label{fig_ablation_numbers}
\end{figure}

\begin{table}
\small
	\centering
	\begin{tabular}{l|c}
		\hline
		  & \text{m-IoU (\%)} \\
		\hline
        (a) without IAF & 62.6\\
        (b) without IAF \& replace attentive pooling & 60.6\\
        (c) without IAF \& replace attentive pooling   & 59.9\\
             \quad \& without correlations & \\
        (d) without multi scale strategy of encoder & 59.2\\
        (e) The Full framework & 64.6\\
		\hline
	\end{tabular}
	\caption{Ablation studies on S3DIS Area 5 validation based on our full network.}
	\label{analysis_ablation_study}
\end{table}


\subsection{Ablation Study}
In this section, we conduct the following ablation studies for our network architecture. All ablated networks are  tested on the Area 5 of S3DIS dataset. Table \ref{analysis_ablation_study} shows the results.

{(a) Removing IAF module.}
This module is used to deal with the indistinguishable points specially. After removing IAF module, we directly feed the output features of feature propagation to the next module.
{(b) Removing IAF module and replacing the attentive pooling with max-pooling.}
The attentive pooling unit learns to automatically combine all local point features in a soft way. By comparison, the max-pooling tends to select or combine features in a hard way, and the performance may be degraded. 
{(c) Based on (b), removing three correlations.}
{(d) Removing multi scale strategy of encoder.}
For enhancing the point's representation of encoder, we use multi scale strategy to obtain features from two different receptive fields. Instead, we use only one receptive field, and the performance is reduced as expected.


\section{Conclusion}
Our paper revolves around the indistinguishable points for semantic segmentation. 
Firstly, we make a qualitative analysis of the indistinguishable points. 
Then we present a novel framework IAF-Net which is based on IAF module and multi-stage loss. 
Besides, we propose a new evaluation metric (IBPM) to evaluate the three types of indistinguishable points respectively.
Experimental results demonstrate the effectiveness and generalization ability of our method.

\section{Acknowledgments}
This work was supported in part by  the Shanghai Committee of Science and Technology, China (Grant No. 20DZ1100800), in part by the National Natural Science Foundation of China under Grant (61876176, U1713208), and in part by the Shenzhen Basic Research Program (CXB201104220032A), Guangzhou Research Program (201803010066).

{

}

\section*{Supplementary Material}

\section*{A.  More Details of IPBM}

\subsection*{A.1 Specific Methods of Subset Point Cloud Acquisition}
\begin{figure}[h]
    \centering
    \includegraphics[width=1\columnwidth]{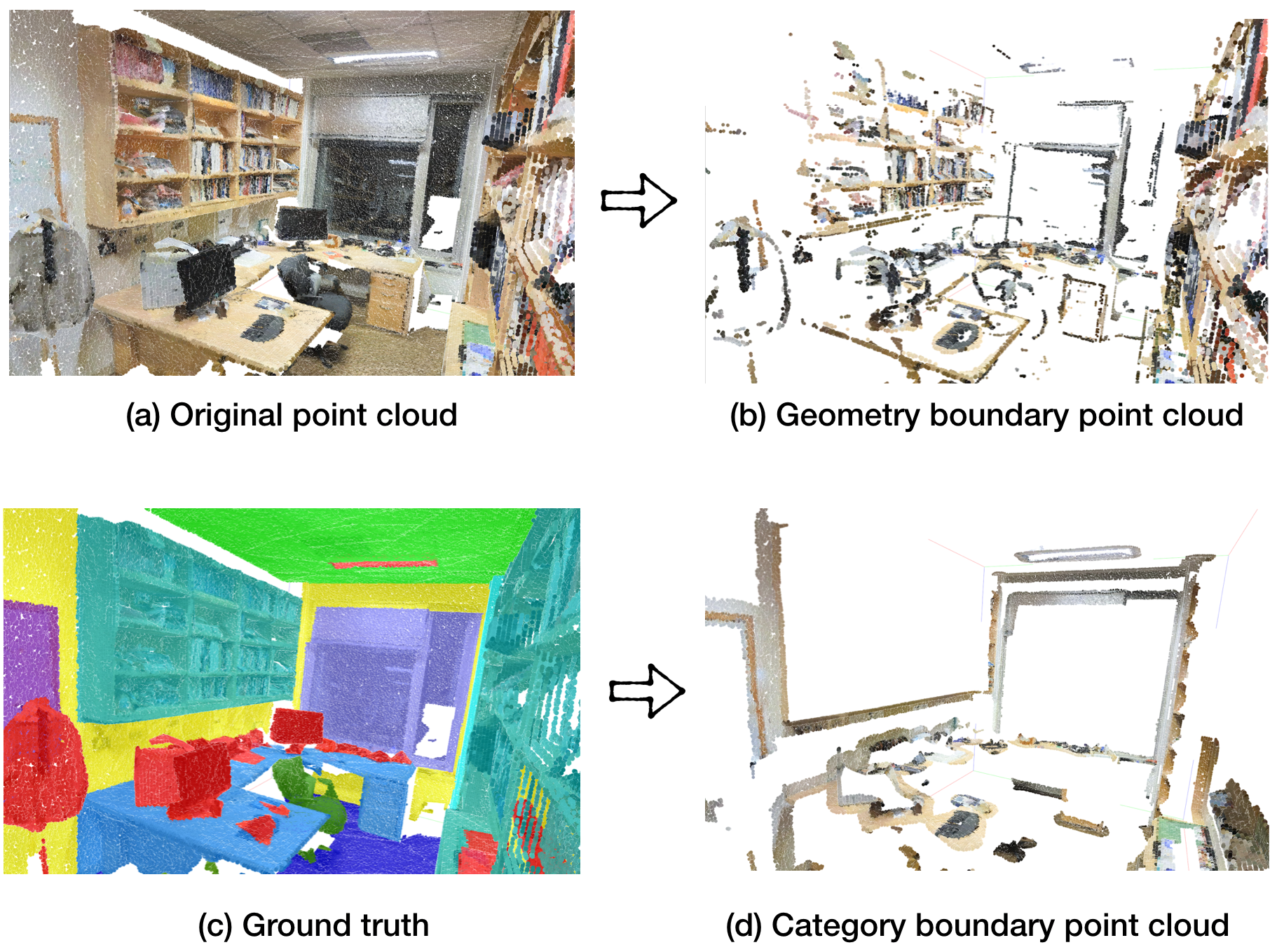} 
    \caption{The process of category boundary and geometry boundary point cloud subset extraction.}
    \label{fig_APP_IPBM_3sub}
\end{figure}

In this section, we will introduce the specific methods of two subset point cloud acquisitions. The two subsets are category boundary point cloud and geometry boundary point cloud (Figure \ref{fig_APP_IPBM_3sub}).
For the whole points $\mathrm{P}=\{p_1,p_2,...,p_N\}$, we have the ground truth labels $Label = \{z_{i,gt}, 1\leq i\leq N \}$. 

\textbf{Category Boundary Point Cloud Acquisition.}
For each point $p_i$, its neighbors in Euclidean space are $\{p_{i_j}, 1\leq j\leq K\}$, we denote the number of neighbor points that satisfy $z_{i,gt} \neq z_{i_k,gt}$ as $r_i$ for each point $p_i$.
Then, we select the points which satisfy $ r_i \geq \rho K $ as the category boundary points. Here we determine $\rho = 0.002$. Figure \ref{fig_APP_IPBM_3sub} (d) shows the category boundary point cloud.

\textbf{Geometry Boundary Point Cloud Acquisition.}
For each point $p_i$, its neighbors in Euclidean space are $\{p_{i_j}, 1\leq j\leq K\}$. 
Firstly, we calculate the local difference of low-level properties (3D coordinates and color information) for each point:
\begin{equation}
    L D_{1}^{l}\left(p_{i}\right)=\sum_{k=1}^{K}\left\|p_{i}-p_{i_{k}}\right\|_{2}
\end{equation}
Then, we align the points in a descending order according to $LD_{1}^{l}$ and choose top $\epsilon N$ points ($\epsilon = 0.25$) as the geometry boundary point cloud which are shown in Figure \ref{fig_APP_IPBM_3sub} (b). 
\begin{figure}[h]
    \centering
    \includegraphics[width=1\columnwidth]{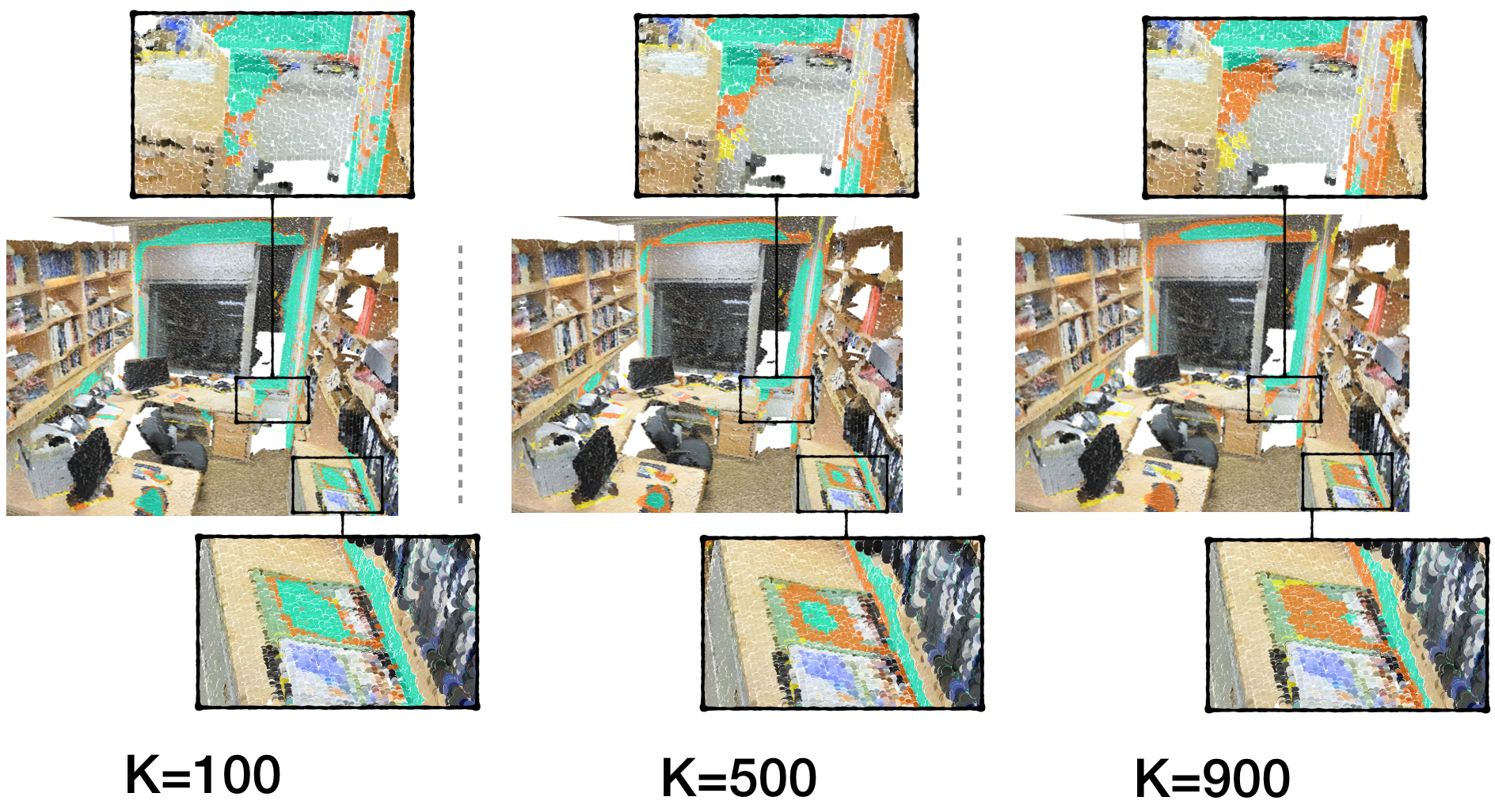} 
    \caption{The visual experiments of K on S3DIS dataset of IPBM. Yellow areas are isolate small areas; Orange areas are complex boundary areas; Cyan areas are confusing interior areas. \textbf{Best viewed in color with 200\% zoom.}
}
    \label{fig_method_IPBM_K}
\end{figure}

\subsection*{A.2 Hyperparameter $K$ in IPBM} 
As shown in Figure \ref{fig_method_IPBM_K}, according to the visualization effect, when $K$ is 500, it is most helpful for us to evaluate the segmentation results in different indistinguishable points areas.
If we use IPBM with fixed K to evaluate the network performance on different datasets, we need to ensure that the density of point cloud is consistent, and FPS can ensure the density consistency.

\subsection*{A.3 Visualization Results of IPBM }
\begin{figure*}[t]
    \centering
    \includegraphics[width=2\columnwidth]{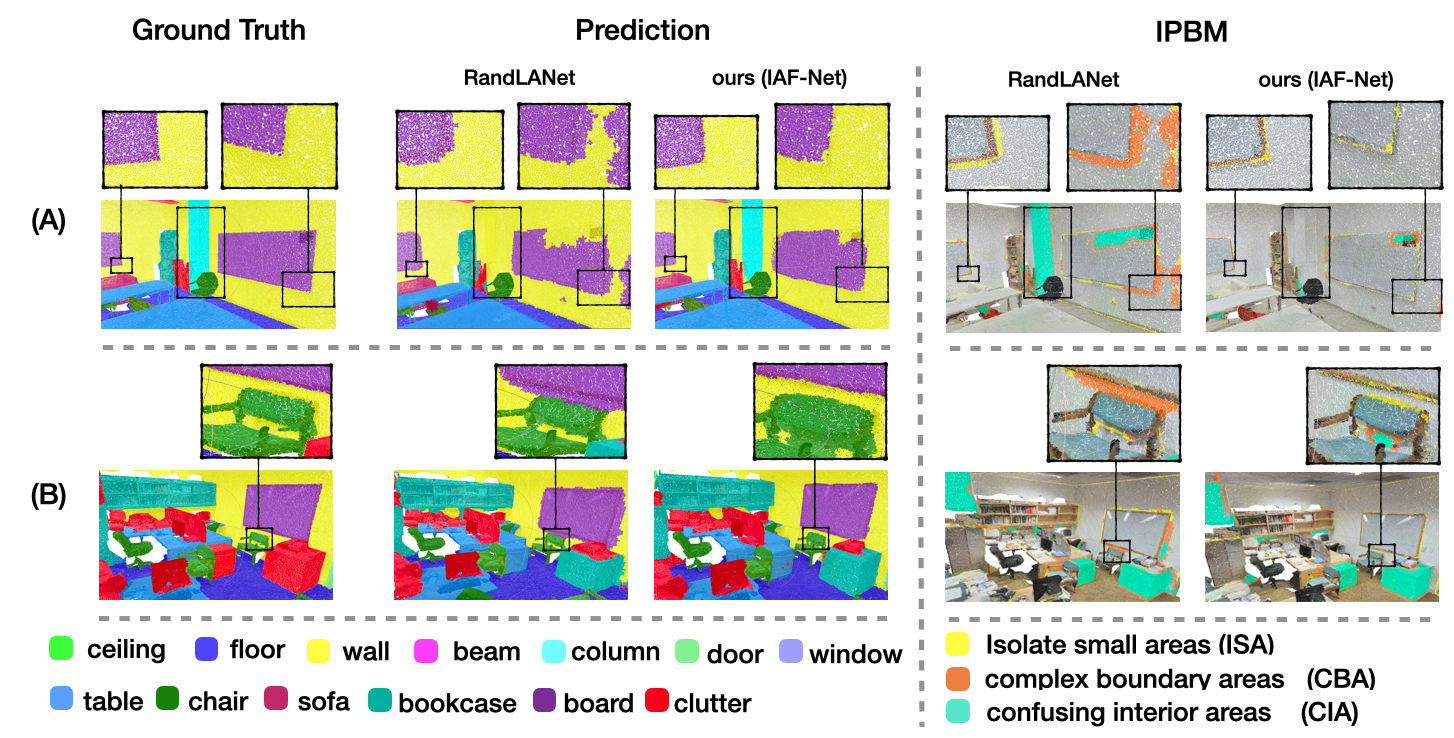} 
    \caption{Visualization Results of prediction and IPBM. (left) are ground truth and predictions of two methods. (right) is the visualization of IPBM. }
    \label{fig_APP_IPBM_result}
\end{figure*}

As shown in Figure \ref{fig_APP_IPBM_result}, m-IOU is capable of showing overall segmentation performance, but it lacks ability to depict the result of segmentation method on specific detail area. Conversely, our IPBM is able to delineate the performances of segmentation methods on different types of indistinguishable areas. Specifically, a higher CIA, such as a cyan-covered column (in Figure \ref{fig_APP_IPBM_result}) whose geometric structure and texture are similar to the wall, represents more misclassified points in confusing interior areas;
a higher CBA, such as orange-covered board edges (in Figure \ref{fig_APP_IPBM_result}) which correspond to category boundaries, represents more misclassified points in complex boundary areas;
a higher ISA, such as yellow-covered small separated spots on wall (in Figure \ref{fig_APP_IPBM_result}), represents more misclassified points in isolated areas. Generally, our new metric is conducive to finding out the specific problems of different methods in semantic segmentation and consequently inspires them to improve methods.

\section*{B. Network Architectures and Parameters}
\subsection*{B.1 Implementation Details of IAF-Net}
Our architecture follows the widely-used encoder-decoder architecture with skip connections which is  shown in Figure 3 in the main paper. The details of each part are as follows:

\noindent\textbf{Network Input:}
The input of our network is a large-scale point cloud with a size of $N \times (3+d)$, where N is the number of points and $(3+d)$ denotes the xyz-dimension and additional properties, such as colors, normal vectors. As for S3DIS dataset \cite{armeni20163d}, each input point is represented by 3D coordinates and color information, while each point of the ScanNet dataset \cite{dai2017scannet} is only represented by the coordinates information.

\noindent\textbf{Network Layers:}
Five encoding layers are utilized in our network to progressively reduce the size of the point clouds and increase the per-point feature dimensions, and four decoding layers are utilized after the encoder. We use Farthest Point Sampling to sample the points. In particular, 25\% points are retrained after each encoding layer, i.e., $\left(N \rightarrow \frac{N}{4} \rightarrow \frac{N}{16} \rightarrow \frac{N}{64} \rightarrow \frac{N}{256}\right)$. Meanwhile, the per-point feature dimension in each layer is $(64 \rightarrow 128 \rightarrow 256 \rightarrow 512 \rightarrow 1024)$.

\begin{figure*}[t]
    \centering
    \includegraphics[width=2\columnwidth]{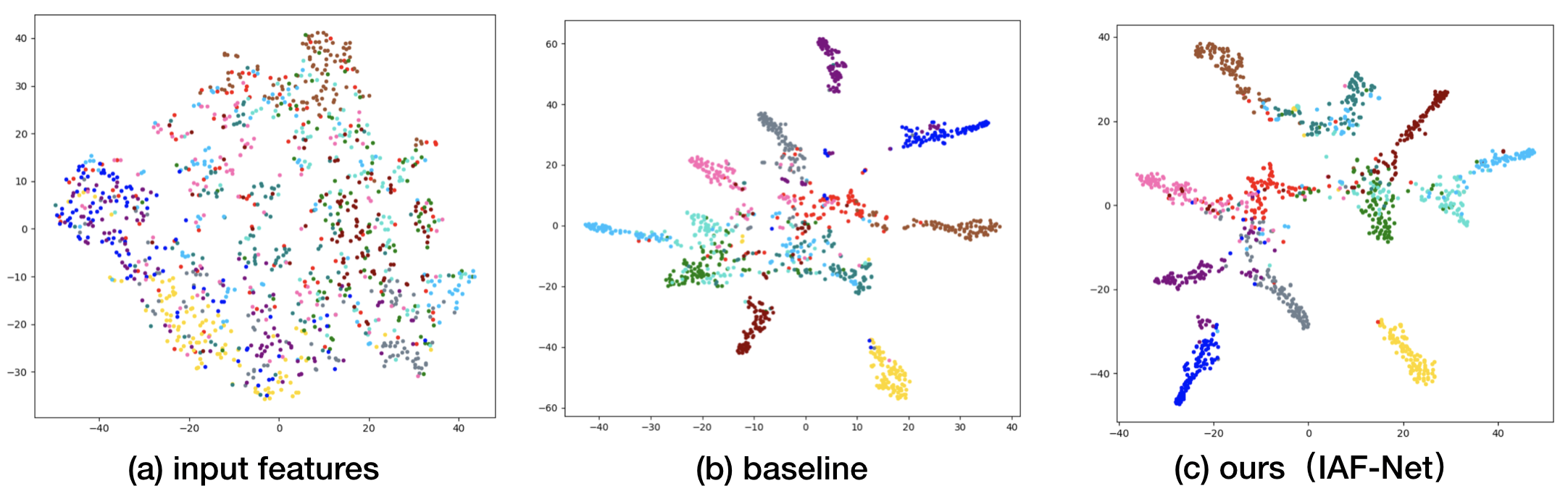} 
    \caption{T-SNE visualization of features on S3DIS datasets (Area 5). (a) input feature, (b) our baseline network's output feature, (c) IAF-Net's output feature.}
    \label{fig_method_tsne}
\end{figure*}

\noindent\textbf{Network Output:}
The final semantic label of each point is obtained through three shared fully-connected layers ($(N, 64) \rightarrow(N, 32) \rightarrow\left(N, {C}\right)$) and a dropout layer with ratio 0.5. The output of our network is the predictions of $N$ points, and the size is $N \times C$, where C is the number of categories.

\subsection*{ B.2 Training Details of IAF-Net}
We conduct our experiments based on the PyTorch \cite{paszke2017automatic} platform on four GTX 2080 GPUs. During the training,
the initial learning rate is 0.001 and a mini-batch size of 8. For the S3DIS dataset, we train for 200 epochs and decay the learning rate by 0.5 for every 20 epochs with Adam optimizer. For the ScanNet, we train for 200 epochs and decay the learning rate by 0.1 for every 50 epochs with SGD optimizer.

\subsection*{B.3 Analysis of Model Complexity}
We analyze the module complexity with our IAF-Net and our baseline. Our baseline is implemented based on GSNet \cite{Xu_2020} which do not include IAF module and multi-stage loss. As shown in Table \ref{analysis_modelSize}, we significantly improve the performance (+2.00\%) while increasing very few parameters (+1.68\%). 
\begin{table}
	\centering
	\begin{tabular}{c|cc}
		\hline
		&  model size  &  mIoU ($\%$)  \\
		\hline
         Baseline  & \textbf{9.30M} & 62.60 \\
         Ours (IAF-Net)  & 10.98M (+1.68$\%$) & \textbf{64.60} (\textbf{+2.00\%}) \\
		\hline
	\end{tabular}
	\caption{Comparisons results on model complexity.}
	\label{analysis_modelSize}
\end{table}
\section*{C. Additional Analysis on IAF-Net}

\begin{table}

	\centering
	\begin{tabular}{c|ccc|c}
		\hline
		&$\mu_1$ & $\mu_2$ & $\mu_3$ & m-IoU ($\%$) \\
		\hline
        (a)& 1 & 0 & 0 & 61.7 \\
        (b)& 0 & 1 & 0 & 62.1 \\
		(c)& 0 & 0 & 1 & 64.6 \\
		(d)& 1/2 & 1/2 & 0 & 62.5 \\
		(e)& 1/3 & 1/3 & 1/3 & 63.7 \\		
		\hline
	\end{tabular}
	\caption{Ablation studies of the three local differences' accumulation.}
	\label{analysis_mu}
\end{table}

\subsection*{C.1 Hyperparameter $\mu$ of the Local Differences Accumulation}
In Section 3.2, \{$LD^l_{1}(P)$,  $LD^l_{2}(P)$, $LD^l_{3}(P)$\} indicate the local difference of original points, semantic predictions and fine-grained features. 
In order to explore the optimal weight factors in the $LD^l(P)$, we conduct experiments on five combinations of $\{ \mu_1, \mu_2, \mu_3 \}$, and the results are shown in Table \ref{analysis_mu}. 
We draw the conclusion from the results in Table \ref{analysis_mu} that utilization of local difference of fine-grained features will improve the performance of our network to an clear extent. To investigate the reason, the local difference of original points ($LD^l_{1}(P)$) only provide coarse information and it is invariant during the training. Although the local difference of semantic predictions $LD^l_{2}(P)$ can mine the indistinguishable points adaptively, 
it can not characterize the features of indistinguishable points in a fine-grained way as $LD^l_{3}(P)$ does. Moreover, $LD^l_{3}(P)$ can also achieve the purpose of adaptive mining according to the fine-grained features.
On S3DIS Area 5 validation, when we use only fine-grained features' local difference (c), we get the best performance. 
On the contrary, when we only use local differences of original points or semantic predictions ((a), (b) in Table \ref{analysis_mu}), we do not achieve the best performance ((c) in Table \ref{analysis_mu}). Then we combine the original points and semantic predictions’ local differences together ((d) in Table \ref{analysis_mu}), the m-IoU is 62.5$\%$. And the combination of three local differences result in 63.7 $\%$ ((e) in Table \ref{analysis_mu}) .

\begin{table}
\small
	\centering
	\begin{tabular}{cccccc|c}
		\hline
	 $L_p$ & $L_{ms}^{1}$ & $L_{ms}^{2}$ & $L_{ms}^{3}$ & $L_{ms}^{4}$ & $L_{ms}^{5}$ & m-IoU $\%$ \\
		\hline
        $\checkmark$ & & & & & & 57.7 \\
       \checkmark & \checkmark & & & & & 60.8 \\
		\checkmark & \checkmark & \checkmark & & & & 61.2 \\
		 \checkmark & \checkmark & \checkmark & \checkmark & & & 61.2 \\
		 \checkmark & \checkmark & \checkmark & \checkmark & \checkmark & & 61.9 \\	
		 \checkmark & \checkmark & \checkmark & \checkmark & \checkmark & \checkmark & 64.6 \\	
		\hline
	\end{tabular}
	\caption{Ablation studies of multi-stage loss.}
	\label{analysis_loss}
\end{table}


\subsection*{C.2 Impact of Multi-stage Loss}
In Section 3.3, we introduce a multi-stage loss to refine the feature descriptions of points progressively in each layer, which assists points of different levels in equipping with more explicit and accurate  semantic representations.  As shown in Table \ref{analysis_loss}, with the gradual accumulation of each layer’s supervision, our network encourages points from global scales to local scales in decoder to form a coherent understanding of the scene, which results in better performance. Meanwhile, thanks to the more explicit semantic features of each point affected by our supervision manner, our IAF module can 
much better as well as more effectively select indistinguishable points and enhance the features of each point.

\subsection*{C.3 Features Visualization}
For full understanding of the our method, we produce the T-SNE  visualization on the input and output features from our baseline network and our IAF-Net. Our baseline is implemented based on GSNet which do not include IAF module and multi-stage loss. We randomly sample 100 points from the full area for each category and plot the feature distribution as illustrated in Figure \ref{fig_method_tsne}. From (a) to (c) are input feature, output feature generated by the baseline network and output feature by our IAF-Net architecture with IAF module. Multi-stage loss is used to enhance the feature of each point in a progressive way.
Compared with the baseline network, points of different categories in IAF-Net are recognized more easily and precisely. The qualitative visualization indicates that the IAF module and multi-stage loss clearly enhance the discriminative ability of point features.



\begin{thebibliography}{36}
\providecommand{\natexlab}[1]{#1}
\providecommand{\url}[1]{\texttt{#1}}
\providecommand{\urlprefix}{URL }
\expandafter\ifx\csname urlstyle\endcsname\relax
  \providecommand{\doi}[1]{doi:\discretionary{}{}{}#1}\else
  \providecommand{\doi}{doi:\discretionary{}{}{}\begingroup
  \urlstyle{rm}\Url}\fi

\bibitem[{Armeni et~al.(2016)Armeni, Sener, Zamir, Jiang, Brilakis, Fischer,
  and Savarese}]{armeni20163d}
Armeni, I.; Sener, O.; Zamir, A.~R.; Jiang, H.; Brilakis, I.; Fischer, M.; and
  Savarese, S. 2016.
\newblock 3d semantic parsing of large-scale indoor spaces.
\newblock In \emph{CVPR}.

\bibitem[{Boulch(2020)}]{boulch2020convpoint}
Boulch, A. 2020.
\newblock ConvPoint: Continuous convolutions for point cloud processing.
\newblock \emph{Computers \& Graphics} .

\bibitem[{Bruna et~al.(2013)Bruna, Zaremba, Szlam, and
  LeCun}]{bruna2013spectral}
Bruna, J.; Zaremba, W.; Szlam, A.; and LeCun, Y. 2013.
\newblock Spectral networks and locally connected networks on graphs.
\newblock \emph{arXiv preprint arXiv:1312.6203} .

\bibitem[{Cao et~al.(2019)Cao, Xu, Lin, Wei, and Hu}]{cao2019gcnet}
Cao, Y.; Xu, J.; Lin, S.; Wei, F.; and Hu, H. 2019.
\newblock Gcnet: Non-local networks meet squeeze-excitation networks and
  beyond.
\newblock In \emph{ICCV Workshops}.

\bibitem[{Charles et~al.(2017)Charles, Su, Kaichun, and Guibas}]{Charles_2017}
Charles, R.~Q.; Su, H.; Kaichun, M.; and Guibas, L.~J. 2017.
\newblock PointNet: Deep Learning on Point Sets for 3D Classification and
  Segmentation.
\newblock \emph{CVPR} .

\bibitem[{Chen et~al.(2017)Chen, Ma, Wan, Li, and Xia}]{chen2017multi}
Chen, X.; Ma, H.; Wan, J.; Li, B.; and Xia, T. 2017.
\newblock Multi-view 3d object detection network for autonomous driving.
\newblock In \emph{CVPR}.

\bibitem[{{Chen} et~al.(2020){Chen}, {Zeng}, {Yang}, {Yu}, {Fu}, and
  {Qu}}]{8805456}
{Chen}, Z.; {Zeng}, W.; {Yang}, Z.; {Yu}, L.; {Fu}, C.~W.; and {Qu}, H. 2020.
\newblock LassoNet: Deep Lasso-Selection of 3D Point Clouds.
\newblock \emph{IEEE Transactions on Visualization and Computer Graphics}
  26(1): 195--204.
\newblock \doi{10.1109/TVCG.2019.2934332}.

\bibitem[{Dai et~al.(2017)Dai, Chang, Savva, Halber, Funkhouser, and
  Nie{\ss}ner}]{dai2017scannet}
Dai, A.; Chang, A.~X.; Savva, M.; Halber, M.; Funkhouser, T.; and Nie{\ss}ner,
  M. 2017.
\newblock Scannet: Richly-annotated 3d reconstructions of indoor scenes.
\newblock In \emph{CVPR}.

\bibitem[{Han et~al.(2020)Han, Wen, Wang, Li, and Li}]{Han_2020}
Han, W.; Wen, C.; Wang, C.; Li, X.; and Li, Q. 2020.
\newblock Point2Node: Correlation Learning of Dynamic-Node for Point Cloud
  Feature Modeling.
\newblock \emph{AAAI} .

\bibitem[{Hornik(1991)}]{hornik1991approximation}
Hornik, K. 1991.
\newblock Approximation capabilities of multilayer feedforward networks.
\newblock \emph{Neural networks} .

\bibitem[{Hu et~al.(2020)Hu, Yang, Xie, Rosa, Guo, Wang, Trigoni, and
  Markham}]{Hu_2020}
Hu, Q.; Yang, B.; Xie, L.; Rosa, S.; Guo, Y.; Wang, Z.; Trigoni, N.; and
  Markham, A. 2020.
\newblock RandLA-Net: Efficient Semantic Segmentation of Large-Scale Point
  Clouds.
\newblock \emph{CVPR} .

\bibitem[{Huang, Wang, and Neumann(2018)}]{huang2018recurrent}
Huang, Q.; Wang, W.; and Neumann, U. 2018.
\newblock Recurrent slice networks for 3d segmentation of point clouds.
\newblock In \emph{CVPR}.

\bibitem[{Jiang et~al.(2018)Jiang, Wu, Zhao, Zhao, and Lu}]{jiang2018pointsift}
Jiang, M.; Wu, Y.; Zhao, T.; Zhao, Z.; and Lu, C. 2018.
\newblock Pointsift: A sift-like network module for 3d point cloud semantic
  segmentation.
\newblock \emph{arXiv preprint arXiv:1807.00652} .

\bibitem[{Komarichev, Zhong, and Hua(2019)}]{Komarichev_2019}
Komarichev, A.; Zhong, Z.; and Hua, J. 2019.
\newblock A-CNN: Annularly Convolutional Neural Networks on Point Clouds.
\newblock \emph{CVPR} .

\bibitem[{Landrieu and Simonovsky(2018)}]{landrieu2018large}
Landrieu, L.; and Simonovsky, M. 2018.
\newblock Large-scale point cloud semantic segmentation with superpoint graphs.
\newblock In \emph{CVPR}.

\bibitem[{Li et~al.(2017)Li, Liu, Luo, Change~Loy, and Tang}]{li2017not}
Li, X.; Liu, Z.; Luo, P.; Change~Loy, C.; and Tang, X. 2017.
\newblock Not all pixels are equal: Difficulty-aware semantic segmentation via
  deep layer cascade.
\newblock In \emph{CVPR}.

\bibitem[{Li et~al.(2018)Li, Bu, Sun, Wu, Di, and Chen}]{li2018pointcnn}
Li, Y.; Bu, R.; Sun, M.; Wu, W.; Di, X.; and Chen, B. 2018.
\newblock Pointcnn: Convolution on x-transformed points.
\newblock In \emph{NIPS}.

\bibitem[{Lin et~al.(2020)Lin, Yan, Huang, Du, Liu, Cui, and
  Han}]{lin2020fpconv}
Lin, Y.; Yan, Z.; Huang, H.; Du, D.; Liu, L.; Cui, S.; and Han, X. 2020.
\newblock FPConv: Learning Local Flattening for Point Convolution.
\newblock In \emph{CVPR}.

\bibitem[{Qi et~al.(2017{\natexlab{a}})Qi, Su, Mo, and Guibas}]{qi2017pointnet}
Qi, C.~R.; Su, H.; Mo, K.; and Guibas, L.~J. 2017{\natexlab{a}}.
\newblock Pointnet: Deep learning on point sets for 3d classification and
  segmentation.
\newblock In \emph{CVPR}.

\bibitem[{Qi et~al.(2017{\natexlab{b}})Qi, Yi, Su, and
  Guibas}]{qi2017pointnet++}
Qi, C.~R.; Yi, L.; Su, H.; and Guibas, L.~J. 2017{\natexlab{b}}.
\newblock Pointnet++: Deep hierarchical feature learning on point sets in a
  metric space.
\newblock In \emph{NIPS}.

\bibitem[{Rusu et~al.(2008)Rusu, Marton, Blodow, Dolha, and
  Beetz}]{rusu2008towards}
Rusu, R.~B.; Marton, Z.~C.; Blodow, N.; Dolha, M.; and Beetz, M. 2008.
\newblock Towards 3D point cloud based object maps for household environments.
\newblock \emph{Robotics and Autonomous Systems} .

\bibitem[{Tatarchenko et~al.(2018)Tatarchenko, Park, Koltun, and
  Zhou}]{tatarchenko2018tangent}
Tatarchenko, M.; Park, J.; Koltun, V.; and Zhou, Q.-Y. 2018.
\newblock Tangent convolutions for dense prediction in 3d.
\newblock In \emph{CVPR}.

\bibitem[{Tchapmi et~al.(2017)Tchapmi, Choy, Armeni, Gwak, and
  Savarese}]{tchapmi2017segcloud}
Tchapmi, L.; Choy, C.; Armeni, I.; Gwak, J.; and Savarese, S. 2017.
\newblock Segcloud: Semantic segmentation of 3d point clouds.
\newblock In \emph{3DV}.

\bibitem[{Thomas et~al.(2019)Thomas, Qi, Deschaud, Marcotegui, Goulette, and
  Guibas}]{thomas2019kpconv}
Thomas, H.; Qi, C.~R.; Deschaud, J.-E.; Marcotegui, B.; Goulette, F.; and
  Guibas, L.~J. 2019.
\newblock KPConv: Flexible and Deformable Convolution for Point Clouds.
\newblock \emph{arXiv preprint arXiv:1904.08889} .

\bibitem[{Wang et~al.(2019)Wang, Huang, Hou, Zhang, and Shan}]{wang2019graph}
Wang, L.; Huang, Y.; Hou, Y.; Zhang, S.; and Shan, J. 2019.
\newblock Graph Attention Convolution for Point Cloud Semantic Segmentation.
\newblock In \emph{CVPR}.

\bibitem[{Wang et~al.(2018{\natexlab{a}})Wang, Suo, Ma, Pokrovsky, and
  Urtasun}]{wang2018deep}
Wang, S.; Suo, S.; Ma, W.-C.; Pokrovsky, A.; and Urtasun, R.
  2018{\natexlab{a}}.
\newblock Deep parametric continuous convolutional neural networks.
\newblock In \emph{CVPR}.

\bibitem[{Wang et~al.(2018{\natexlab{b}})Wang, Girshick, Gupta, and
  He}]{Wang_2018}
Wang, X.; Girshick, R.; Gupta, A.; and He, K. 2018{\natexlab{b}}.
\newblock Non-local Neural Networks.
\newblock \emph{CVPR} .

\bibitem[{Wang et~al.(2018{\natexlab{c}})Wang, Sun, Liu, Sarma, Bronstein, and
  Solomon}]{wang2018dynamic}
Wang, Y.; Sun, Y.; Liu, Z.; Sarma, S.~E.; Bronstein, M.~M.; and Solomon, J.~M.
  2018{\natexlab{c}}.
\newblock Dynamic graph cnn for learning on point clouds.
\newblock \emph{arXiv preprint arXiv:1801.07829} .

\bibitem[{Wu, Qi, and Fuxin(2019)}]{wu2019pointconv}
Wu, W.; Qi, Z.; and Fuxin, L. 2019.
\newblock Pointconv: Deep convolutional networks on 3d point clouds.
\newblock In \emph{CVPR}.

\bibitem[{Xu, Zhou, and Qiao(2020)}]{Xu_2020}
Xu, M.; Zhou, Z.; and Qiao, Y. 2020.
\newblock Geometry Sharing Network for 3D Point Cloud Classification and
  Segmentation.
\newblock \emph{AAAI} .

\bibitem[{Yan et~al.(2020)Yan, Zheng, Li, Wang, and Cui}]{yan2020pointasnl}
Yan, X.; Zheng, C.; Li, Z.; Wang, S.; and Cui, S. 2020.
\newblock PointASNL: Robust Point Clouds Processing using Nonlocal Neural
  Networks with Adaptive Sampling.
\newblock In \emph{CVPR}.

\bibitem[{Yang et~al.(2019)Yang, Zhang, Ni, Li, Liu, Zhou, and
  Tian}]{yang2019modeling}
Yang, J.; Zhang, Q.; Ni, B.; Li, L.; Liu, J.; Zhou, M.; and Tian, Q. 2019.
\newblock Modeling point clouds with self-attention and gumbel subset sampling.
\newblock In \emph{CVPR}.

\bibitem[{Ye et~al.(2018)Ye, Li, Huang, Du, and Zhang}]{ye20183d}
Ye, X.; Li, J.; Huang, H.; Du, L.; and Zhang, X. 2018.
\newblock 3d recurrent neural networks with context fusion for point cloud
  semantic segmentation.
\newblock In \emph{ECCV}.

\bibitem[{Zhang, Luo, and Urtasun(2018)}]{zhang2018efficient}
Zhang, C.; Luo, W.; and Urtasun, R. 2018.
\newblock Efficient convolutions for real-time semantic segmentation of 3d
  point clouds.
\newblock In \emph{3DV}.

\bibitem[{Zhang, Hua, and Yeung(2019)}]{zhang-shellnet-iccv19}
Zhang, Z.; Hua, B.-S.; and Yeung, S.-K. 2019.
\newblock ShellNet: Efficient Point Cloud Convolutional Neural Networks using
  Concentric Shells Statistics.
\newblock In \emph{ICCV}.

\bibitem[{Zhao et~al.(2019)Zhao, Jiang, Fu, and Jia}]{zhao2019pointweb}
Zhao, H.; Jiang, L.; Fu, C.-W.; and Jia, J. 2019.
\newblock PointWeb: Enhancing local neighborhood features for point cloud
  processing.
\newblock In \emph{CVPR}.

\end{thebibliography}
\end{document}